\documentclass[12pt, final]{cls/l4dc2023}

\title[Lie Group Forced Variational Integrator Networks]{Lie Group Forced Variational Integrator Networks\\for Learning and Control of Robot Systems}
\usepackage{times}

\usepackage{amsfonts,amscd,dsfont} 
\usepackage{graphicx,wrapfig,adjustbox}
\usepackage{times}

\usepackage{dutchcal}

\newcommand{\bfu}{\mathbf{u}}

\newcommand{\bftau}{{\boldsymbol{\tau}}}

\newcommand{\bbR}{\mathbb{R}}

\newtheorem*{proposition*}{Proposition}
\newtheorem*{corollary*}{Corollary}

\newtheorem*{assumption*}{Assumption}
\newtheorem*{problem*}{Problem}
\newtheorem{problem}{Problem}
\newtheorem*{solution*}{Solution}
\newtheorem*{remark*}{Remark}

\definecolor{mygreen}{rgb}{0, 0.5, 0}

\author{%
 \Name{Valentin Duruisseaux} \Email{vduruiss@ucsd.edu}\\
 \addr Department of Mathematics, University of California San Diego, La Jolla, CA 92093
 \AND
 \Name{Thai Duong} \Email{tduong@ucsd.edu}\\
 \addr Department of Electrical and Computer Engineering, University of California San Diego, La Jolla, CA~92093\AND
 \Name{Melvin Leok} \Email{mleok@ucsd.edu}\\
 \addr Department of Mathematics, University of California San Diego, La Jolla, CA 92093
 \AND
 \Name{Nikolay Atanasov} \Email{natanasov@ucsd.edu}\\
 \addr Department of Electrical and Computer Engineering, University of California San Diego, La Jolla, CA~92093}

\begin{document}

\maketitle

\begin{abstract}%
Incorporating prior knowledge of physics laws and structural properties of dynamical systems into the design of deep learning architectures has proven to be a powerful technique for improving their computational efficiency and generalization capacity. Learning accurate models of robot dynamics is critical for safe and stable control. Autonomous mobile robots, including wheeled, aerial, and underwater vehicles, can be modeled as controlled Lagrangian or Hamiltonian rigid-body systems evolving on matrix Lie groups. In this paper, we introduce a new structure-preserving deep learning architecture, the Lie group Forced Variational Integrator Network (LieFVIN), capable of learning controlled Lagrangian or Hamiltonian dynamics on Lie groups, either from position-velocity or position-only data. By design, LieFVINs preserve both the Lie group structure on which the dynamics evolve and the symplectic structure underlying the Hamiltonian or Lagrangian systems of interest. The proposed architecture learns surrogate discrete-time flow maps allowing accurate and fast prediction without numerical-integrator, neural-ODE, or adjoint techniques, which are needed for vector fields. Furthermore, the learnt discrete-time dynamics can be utilized with computationally scalable discrete-time (optimal) control strategies. 
\end{abstract}

\begin{keywords}%
Dynamics Learning, Variational Integrators, Symplectic Integrators, Structure-Preserving Neural Networks, Physics-Informed Machine Learning, Predictive Control, Lie Group Dynamics
\end{keywords}


\section{Introduction} \label{sec:intro}

Dynamical systems evolve according to physics laws which can be described using differential equations. An accurate model of the dynamics of a control system is important, not only for predicting its future behavior, but also for designing control laws that ensure desirable properties such as safety, stability, and generalization to different operational conditions. 

This paper considers the problem of learning dynamics: given a dataset of trajectories from a dynamical system, we wish to infer the update map that generates these trajectories and use it to predict the evolution of the system from different initial states. Models obtained from first principles are used extensively in practice but tend to over-simplify the underlying structure of dynamical systems, leading to prediction errors that cannot be corrected by optimizing over a few model parameters. Deep learning provides very expressive models for function approximation but standard neural networks struggle to learn the symmetries and conservation laws underlying dynamical systems, and as a result do not generalize well. Deep learning models capable of learning and generalizing dynamics effectively \citep{Willard2020} are typically over-parameterized and require large datasets and substantial training time, making them prohibitively expensive for applications such as robotics.

A recent research direction has been considering a hybrid approach, which encodes physical laws and geometric properties of the underlying system in the design of the neural network architecture or in the learning process. Prior physics knowledge can be used to construct physics-informed neural networks with improved design and efficiency and better generalization capacity, which take advantage of the function approximation power of neural networks to handle incomplete knowledge. In this paper, we consider learning controlled Lagrangian or Hamiltonian dynamics on Lie groups while preserving the symplectic structure underlying these systems and the Lie group constraints. 

Symplectic maps possess numerous special properties and are closely related to Hamiltonian systems. Preserving the symplectic structure of a Hamiltonian system when constructing a discrete approximation of its flow map ensures the preservation of many aspects of the system such as total energy, and leads to physically well-behaved discrete solutions~\citep{LeRe2005,HaLuWa2006,Holm2009,Blanes2017}. It is thus important to have structure-preserving architectures which can learn flow maps and ensure that the learnt maps are symplectic. Many physics-informed approaches have recently been proposed to learn Hamiltonian dynamics and symplectic maps \citep{Lutter2018,Greydanus2019,Bertalan2019,Jin2020,BurbyHenon,Chen2020,Cranmer2020,Zhong2020,zhong2020dissipative,Zhong2021,Marco2021,Rath2021,Chen2021neural,Offen2022,Santos2022,Valperga2022,Mathiesen2022,DuruisseauxNPMap}.

Our physics-informed strategy, inspired by (Forced) Variational Integrator Networks ((F)VINs) \citep{Saemundsson2020,Havens2021}, differs from most of these approaches by learning a discrete-time symplectic approximation to the flow map of the dynamical system, instead of learning the vector field for the continuous-time dynamics. This allows fast prediction for simulation, planning and control without the need to integrate differential equations or use neural ODEs and adjoint techniques. Additionally, the learnt discrete-time dynamics can be combined with computationally scalable discrete-time control strategies. 

The novelty of our approach with respect to (F)VINs resides in the enforcement not only of the preservation of symplecticity but also of the Lie group structure when learning a surrogate map for a controlled Lagrangian system which evolves on a Lie group. This is achieved by working in Lie group coordinates instead of Euclidean coordinates, by matching the training data to a parameterized forced Lie group variational integrator which evolves intrinsically on the Lie group. More specifically, we extend the discrete-time Euclidean formulation of FVINs with control from \citep{Havens2021} to Lie groups in a structure-preserving way, which is particularly relevant when considering robot systems (e.g., wheeled, aerial, and underwater vehicles) since they can often be modeled as controlled Lagrangian rigid-body systems evolving on Lie groups.

Given a learnt dynamical system, it is often desirable to control its behavior to achieve stabilization, tracking, or other control objectives. Control designs for continuous-time Hamiltonian systems rely on the Hamiltonian structure \citep{lutter2019deepunderactuated,Zhong2020, duong21hamiltonian, duong2022adaptive}. Since the Hamiltonian captures the system energy, control techniques for stabilization inject additional energy into the system via the control input to ensure that the minimum of the total energy is at a desired equilibrium. For fully-actuated Hamiltonian systems, it is sufficient to shape the potential energy only using energy-shaping and damping-injection (ES-DI) \citep{van2014port}. For under-actuated systems, both the kinetic and potential energies are shaped, e.g., via interconnection and damping assignment passivity-based control (IDA-PBC) \citep{ortega2002stabilization,van2014port,acosta2014robust,cieza2019ida}. The most widely used control approach for discrete-time dynamics is based on Model Predictive Control (MPC) \citep{borrelli_MPC_book,NMPC_book}. MPC techniques determine an open-loop control sequence that solves a finite-horizon optimal control problem, apply the first few control inputs, and repeat the process. A key result in MPC is that an appropriate choice of terminal cost and terminal constraints in the sequence of finite-horizon problems can guarantee recursive feasibility and asymptotic optimality with respect to the infinite-horizon cost \citep{borrelli_MPC_book}. The ability to learn a structure-preserving discrete-time model of a dynamics system enabled by this paper, also allows employing MPC techniques for optimal control of the learnt system dynamics.

\section{Preliminaries}
\label{sec:prelim}

We first review the basic theory of continuous-time Lagrangian and Hamiltonian systems, before describing their underlying symplectic structure and how variational integrators preserve that structure. Finally, we discuss how external forcing and control can be added to variational integrators.

\subsection{Geometric Mechanics}
The set of tangent vectors to a manifold $\mathcal{Q} $ at a point $q\in \mathcal{Q}$ is a vector space called the tangent space $T_q \mathcal{Q}$ to $\mathcal{Q}$ at $q$. The disjoint union of all the tangent spaces to $\mathcal{Q}$ forms the tangent bundle $T\mathcal{Q} = \{  (q,v) | q\in \mathcal{Q}, v \in T_q\mathcal{Q}  \}$ of $\mathcal{Q}$.  The vector space dual to the tangent space $T_q \mathcal{Q}$ is the cotangent space $T_q^* \mathcal{Q}$, and the vector bundle over $\mathcal{Q}$ whose fibers are the cotangent spaces of $\mathcal{Q}$ is the cotangent bundle $T^*\mathcal{Q}  = \{  (q,p) | q\in \mathcal{Q}, p \in T^*_q \mathcal{Q} \}$.

Given a manifold $\mathcal{Q} $, a Lagrangian is a function $L:T\mathcal{Q}  \rightarrow \mathbb{R}$. Hamilton's Variational Principle states that $ \delta \int_{0}^{T}{L(q(t),\dot{q}(t))dt} = 0$, where the variation is induced by an infinitesimal variation $\delta q$ that vanishes at the endpoints. Hamilton's Principle is equivalent to the Euler--Lagrange equations
\begin{equation}\label{eq: EL Basic}
 \frac{\partial L }{\partial q}(q,\dot{q}) -	\frac{d}{dt} \left( \frac{\partial L} {\partial \dot{q}} (q,\dot{q}) \right) = 0.
\end{equation}
Given a Lagrangian $L$, we define the conjugate momentum $p \in T^* \mathcal{Q}$ via the Legendre transform $p = \frac{\partial L}{\partial \dot{q}}$, and obtain a Hamiltonian  $H(q,p) = \sum_{j=1}^{n} {p_j  \dot{q}^j} - L(q,\dot{q})\big|_{p_i=\frac{\partial L}{\partial \dot{q}^i}}$ on $T^*\mathcal{Q}$. There is a variational principle on the Hamiltonian side which is equivalent to Hamilton's equations and to the Euler--Lagrange equations~\eqref{eq: EL Basic} when the Legendre transform is diffeomorphic. For most mechanical systems, the Legendre transform is diffeomorphic and thus the Lagrangian and Hamiltonian formulations are equivalent. The approaches presented here are based on the Lagrangian formulation, but also apply to the equivalent Hamiltonian systems whenever they are well-defined.

\subsection{Symplecticity}

A smooth mapping $(q,p) \mapsto (\bar{q},\bar{p})$ is symplectic if it preserves the symplectic two-form, that is $\sum_{i=1} \mathbf{d}q^i\wedge \mathbf{d}p_i = \sum_{i=1} \mathbf{d}\bar{q}^i\wedge \mathbf{d}\bar{p}_i$. Hamiltonian systems and symplectic flows are closely related: solutions to Hamiltonian systems are symplectic flows \citep{Poincare1899}, and symplectic flows are locally Hamiltonian. When applied to Hamiltonian systems, symplectic integrators yield discrete approximations of the flow that preserve the symplectic two-form, which results in the preservation of many qualitative aspects of the dynamical system and leads to physically well-behaved solutions. See \citep{LeRe2005,HaLuWa2006,Blanes2017} for a comprehensive presentation of geometric numerical integration.

\subsection{Variational Integrators}

 Variational integrators are obtained by discretizing Hamilton's principle, instead of discretizing the equations of motion, are thus symplectic, preserve many invariants, and exhibit excellent long-time near-energy preservation~\citep{MaWe2001}. The exact discrete Lagrangian generating the time-$h$ flow can be represented in boundary-value form as $L_d^E(q_0,q_1)=\int_0^h L(q(t),\dot q(t)) dt ,$ where $q(t)$ satisfies the Euler--Lagrange equations on $[0,h]$ with $q(0)=q_0$, $q(h)=q_1$. After constructing an approximation $L_d$ to $L_d^E$, the Lagrangian variational integrator is defined implicitly by the discrete Euler--Lagrange equation, $D_2 L_d(q_{k-1} , q_k) + D_1 L_d(q_k, q_{k+1})  = 0,$ which can also be written in Hamiltonian form, using discrete momenta $p_k$, as $
	p_k=-D_1 L_d(q_k, q_{k+1})$ and $ p_{k+1}=D_2 L_d(q_k, q_{k+1})$,  
where $D_i$ denotes a partial derivative with respect to the $i$-th argument. Many properties of the integrator, such as momentum conservation and error analysis guarantees, can be determined by analyzing the discrete Lagrangian, instead of analyzing the integrator directly. 

Examples of variational integrators include Taylor~\citep{ScShLe2017}, Galerkin~\citep{MaWe2001,LeZh2011}, prolongation-collocation~\citep{LeSh2011}, and constrained~\citep{MaWe2001,Duruisseaux2022Constrained} variational integrators. Variational integrators can also be developed for Hamiltonian dynamics \citep{LaWe2006, LeZh2011,ScLe2017,duruisseaux2020adaptive}, and can be used with prescribed variable time-steps \citep{duruisseaux2020adaptive,Duruisseaux2022Lagrangian}.

\subsection{Forced Variational Integrators}

External forcing and control can be added to variational integrators \citep{MaWe2001,Ober2011}. Let $u(t)$ be the control parameter in some control manifold $\mathcal{U}$, and consider a Lagrangian control force $\mathcal{f}_L : T\mathcal{Q} \times \mathcal{U} \rightarrow T^* \mathcal{Q}$. Hamilton's principle can be modified into the Lagrange--d'Alembert Principle
\begin{equation} \label{eq: Lagrange dAlembert}
    \delta \int_{0}^{T}{L(q(t),\dot{q}(t))dt} + \int_0^T{\mathcal{f}_L(q(t),\dot{q}(t),u(t)) \cdot \delta q(t) dt} =0,
\end{equation} 
where the variation is induced by an infinitesimal variation $\delta q$ that vanishes at the endpoints. This variational principle is equivalent to the forced Euler--Lagrange equations
\begin{equation}\label{eq: EL Forced}
 \frac{\partial L }{\partial q}(q,\dot{q}) -	\frac{d}{dt} \left( \frac{\partial L} {\partial \dot{q}} (q,\dot{q}) \right) + \mathcal{f}_L(q,\dot{q},u) = 0.
\end{equation}
Using a discrete Lagrangian $L_d$, and discrete Lagrangian control forces $\mathcal{f}_d^{\pm} : \mathcal{Q} \times \mathcal{Q} \times \mathcal{U} \rightarrow T^* \mathcal{Q}$ to approximate the virtual work of the Lagrangian control force $\mathcal{f}_L$,
\begin{equation}
    \int_{t_k}^{t_{k+1}}{\mathcal{f}_L(q(t),\dot{q}(t),u(t)) \cdot \delta q(t) dt} \approx  \mathcal{f}_d^-(q_k,q_{k+1},u_k) \cdot \delta q_k +  \mathcal{f}_d^+(q_k,q_{k+1},u_k)\cdot \delta q_{k+1},
\end{equation}
one can obtain a forced variational integrator from the forced discrete Euler--Lagrange equations
\begin{equation}
	   D_2 L_d(q_{k-1} , q_k)  + D_1 L_d(q_k, q_{k+1}) + \mathcal{f}_d^{+}(q_{k-1},q_k, u_{k-1}) + \mathcal{f}_d^{-}(q_k, q_{k+1},u_k) = 0,
	\end{equation}
which can also be written in Hamiltonian form as
\begin{align}
	p_k  =-D_1 L_d(q_k, q_{k+1}) - \mathcal{f}_d^-(q_k,q_{k+1},u_k), \quad  p_{k+1} =D_2 L_d(q_k, q_{k+1}) + \mathcal{f}_d^+(q_k,q_{k+1},u_k). \label{IDEL-Forced}
\end{align}

\section{Problem Statement}
\label{sec: problem statement}

We consider the problem of learning controlled Lagrangian dynamics. Given a position-velocity dataset of trajectories, we wish to infer the flow map generating these trajectories, while preserving the system's symplectic structure and constraining the updates to the Lie group on which it evolves. For example, a rigid-body robot system may be modeled as a Lagrangian system evolving on the Lie group $\text{SE}(3)$ of rigid-body transformations. Learning its dynamics from trajectory data should respect kinematic and energy conservation. More precisely, we consider the following problem. \vspace{-2mm}

\begin{problem} \label{problem1}
 Let $\mathcal{Q}$
 be a Lie group and $\mathcal{D}_{T \mathcal{Q}}$ be a distance metric on $T \mathcal{Q}$. Given a dataset of position-velocity updates $\left\{ \left(q_0^{(i)}, \dot{q}_0^{(i)}, u_0^{(i)}\right) \mapsto \left(q_1^{(i)}, \dot{q}_1^{(i)}\right)  \right\}_{i=1}^{N}$ for a controlled Lagrangian dynamical system evolving on $\mathcal{Q}$, we wish to find a symplectic mapping $\Psi : T \mathcal{Q}  \times \mathcal{U} \rightarrow T\mathcal{Q} $ which minimizes  \vspace{-0.5mm}
 \begin{equation}
     \sum_{i=1}^{N}{\mathcal{D}_{T \mathcal{Q}}\left(\left(q_1^{(i)}, \dot{q}_1^{(i)}\right) , \Psi\left(q_0^{(i)}, \dot{q}_0^{(i)}, u_0^{(i)}\right) \right)}.  \vspace{-0.4mm}
 \end{equation}

\end{problem}

\section{Lie group Forced Variational Integrators Networks (LieFVINs)}
\label{sec:technical_approach}

To solve Problem \ref{problem1}, we introduce Lie group Forced Variational Integrators Networks (\textcolor{blue}{\textbf{LieFVINs}}). Our main idea is to parametrize the updates of a forced Lie group variational integrator and match them with observed updates. We focus on specific forced $\text{SO}(3)$ and $\text{SE}(3)$ variational integrators, but the general strategy extends to any Lie group forced variational integrator.

\subsection{The $\text{SO}(3)$ and $\text{SE}(3)$ Lie Groups}

The 3-dimensional special orthogonal group $\text{SO}(3) = \{ R \in \mathbb{R}^{3\times 3} | RR^\top = \mathbb{I}_3, \det{(R)} = 1 \}$, where $\mathbb{I}_k $ denotes the $k \times k$ identity matrix, is the Lie group of rotations about the origin in $\mathbb{R}^3$. The Lie algebra  of $\text{SO}(3)$ is the space of skew-symmetric matrices $\mathfrak{so}(3) = \{ A \in \mathbb{R}^{3\times 3} |  A^\top  = -A \},$
with the matrix commutator $[A,B] = AB-BA$ as the Lie bracket. The sets $\mathbb{R}^{3} $ and $ \mathfrak{so}(3)$ are isomorphic via the hat map $S(\cdot) : \mathbb{R}^{3} \rightarrow \mathfrak{so}(3) $, defined by $S(x) y = x\times y$ for any $ x,y\in \mathbb{R}^3$. 

The Special Euclidean group in 3 dimensions, $\text{SE}(3)$, is a semidirect product of $\mathbb{R}^3$ and $\text{SO}(3)$ and is diffeomorphic to $\mathbb{R}^3 \times \text{SO}(3)$. Elements of $\text{SE}(3)$ can be written as $(x,R) \in \mathbb{R}^3 \times \text{SO}(3)$, and the Lie algebra $\mathfrak{se}(3)$ of $\text{SE}(3)$ is composed of elements $(y,A) \in \mathbb{R}^3 \times \mathfrak{so}(3)$. 

The pose of a rigid body can be described by an element $(x,R)$ of $\text{SE}(3)$, consisting of position $x \in \bbR^3$ and orientation $R \in \text{SO}(3)$. See Appendix~\ref{appendix: Rigid-Body Kinematics} for more details about rigid-body kinematics.

\subsection{Forced Variational Integrator on $\text{SO}(3)$ and $\text{SE}(3)$}  \label{sec: Variational Integrator}

On $\text{SE}(3)$, $q = (x,R)$ and $\dot{q} = (v,\omega)$ where $x$ is position, $R$ is orientation, $v$ is velocity, and $\omega$ is angular velocity. A Lagrangian on $\text{SE}(3)$ is given by \vspace{-1mm}
\begin{equation} 
L(x,R,v,\omega) = \frac{1}{2} v^\top mv  +\frac{1}{2} \omega^\top  J \omega  - U(x,R),  
\end{equation}  
where $m$ is mass, $J \in \mathbb{R}^{3\times 3}$ is a symmetric positive-definite inertia matrix, $U$ is potential energy. \\

Consider the continuous-time kinematics equation $\dot{R} = RS(\omega)$, with constant $\omega(t) \equiv \omega_k$ for a short period of time $t \in [t_k,t_{k+1})$ where $t_{k+1} = t_k + h$. Then, $R(t_{k+1}) = R(t_k)\exp(h S(\omega_k))$. Thus, with $R_k := R(t_k)$, $R_{k+1} := R(t_{k+1})$ and $Z_k := \exp(h S(\omega_k))$, we obtain $R_{k+1} = R_k Z_k$ and for sufficiently small $h$, we have $Z_k \approx \mathbb{I}_3 + h S(\omega_k)$. With $(x_k,R_k) \in \text{SE}(3)$, the discrete $\text{SE}(3)$ kinematic equations are given by $R_{k+1} = R_k Z_k$ and $x_{k+1} = x_k +R_k y_k$ where $ (y_k, Z_k)\in \text{SE}(3)$, which ensures that the sequence of updates $\{(x_k, R_k) \}_k$ remains on $\text{SE}(3)$. 

Using the approximation $S(\omega_k) \approx \frac{1}{h} ( Z_k - \mathbb{I}_3 )$, we choose the discrete Lagrangian
\begin{equation}  \label{eq: discrete Lagrangian}
\begin{aligned}
L_d(x_k, R_k, y_k, Z_k  )  &  =  \frac{m}{2h} y_k^\top y_k + \frac{1}{h} \text{tr}\left( [\mathbb{I}_3  - Z_k] J_d \right)  \\ &  \qquad \quad - (1-\alpha ) h U(x_k, R_k) - \alpha  h  U(x_k + R_ky_k, R_k Z_k),
\end{aligned}
\end{equation} 
where $\alpha \in [0,1]$ and $J_d = \frac{1}{2} \text{tr}(J) \mathbb{I}_3 - J$. Equation~\eqref{eq: discrete Lagrangian} gives a simple approximation to the exact $\text{SE}(3)$ discrete Lagrangian, while maintaining some flexibility in the two-point quadrature weights through the tunable parameter $\alpha$. Higher-order approximations could also be used, but the resulting discrete equations of motion would typically be more complicated and expensive to evolve. 

We denote $U_k = U(x_k,R_k)$ and define $\xi_k$ via $ S(\xi_k) = \frac{\partial  U_{k}}{\partial R_{k}}^\top R_{k} - R_{k}^\top \frac{\partial U_{k}}{\partial R_{k}}$. In Appendix~\ref{appendix: Derivation of integrator}, we show that the forced discrete Euler--Lagrange equations associated to the discrete Lagrangian~\eqref{eq: discrete Lagrangian} and discrete control forces $\mathcal{f}_{d_k}^{\pm} \equiv \mathcal{f}_{d}^{\pm}(x_k, R_k, u_k)$ with $R$ and $x$ components $\mathcal{f}_{d}^{R\pm},\mathcal{f}_{d}^{x\pm}$ can be written in Hamiltonian form, using $\pi_k = J \omega_k  $ and $\gamma_k = m v_k $, as
\begin{align}
   & hS(\pi_k)   +   h S( \mathcal{f}_{d_k}^{R-}) + (1-\alpha)h^2 S(\xi_k)  =  Z_k J_d - J_d Z_k^\top    \label{eq: SE3 discrete H equation 2}, \\ &
    R_{k+1}  = R_k Z_k  \label{eq: SE3 discrete H equation 5} , \\
    & \pi_{k+1}  = Z_k^\top \pi_k + (1-\alpha)h Z_k^\top \xi_k+ \alpha h \xi_{k+1}   + Z_k^\top \mathcal{f}_{d_k}^{R-}+ \mathcal{f}_{d_k}^{R+}   \label{eq: SE3 discrete H equation 1}, \\ &
      x_{k+1}  =  x_{k} + \frac{h}{m} \gamma_k - (1-\alpha)\frac{h^2}{m} \frac{\partial U_k}{\partial x_k} -  \frac{h}{m} R_k \mathcal{f}_{d_k}^{x-}  \label{eq: SE3 discrete H equation 3}, \\ &
        \gamma_{k+1}  = \gamma_k - (1-\alpha)h    \frac{\partial U_k}{\partial x_k} - \alpha h \frac{\partial U_{k+1}}{\partial x_{k+1}} +  R_k \mathcal{f}_{d_k}^{x-} + R_{k+1} \mathcal{f}_{d_k}^{x+}   \label{eq: SE3 discrete H equation 4}.
\end{align}
Given $(x_k,R_k,\gamma_k, \pi_k, u_k)$, we first solve equation \eqref{eq: SE3 discrete H equation 2} which is of the form $S(a) = Z J_d  - J_d Z^\top$ as outlined in Remark~\ref{remark: Cayley}, and then get $R_{k+1} = R_k Z_k$. We then obtain $\pi_{k+1}$,  $x_{k+1}$ and $\gamma_{k+1}$ from equations~\eqref{eq: SE3 discrete H equation 1}-\eqref{eq: SE3 discrete H equation 4}. The discrete equations of motion can be rewritten as an update from $(x_k,  R_k, v_k,\omega_k, u_k)$ to $(x_{k+1}, R_{k+1}, v_{k+1}, \omega_{k+1})$ by using $\pi_k = J \omega_k  $ and $\gamma_k = m v_k $.

\begin{remark}\label{remark: Cayley}  $ S(a) = Z J_d  - J_d Z^\top$ can be converted into an equivalent vector equation 
\begin{equation} \label{eq: Vector Equation General Form}
   \phi(\mathcal{z}) \equiv a + a \times \mathcal{z} + \mathcal{z}(a^\top \mathcal{z}) - 2J\mathcal{z} = 0, \qquad  \mathcal{z}\in \mathbb{R}^3,
\end{equation}
as shown in Appendix~\ref{appendix: Equation Transformation}, using the Cayley transform 
\begin{equation} Z = \text{Cay}(\mathcal{z}) \equiv  (\mathbb{I}_3+S(\mathcal{z}))(\mathbb{I}_3-S(\mathcal{z}))^{-1} = \frac{1}{1+ \| \mathcal{z} \|_2^2}\left((1-\| \mathcal{z} \|_2^2) \mathbb{I}_3 + 2 S(\mathcal{z}) + 2 \mathcal{z} \mathcal{z}^\top \right). \vspace{-0.55mm} \end{equation}

The solution $Z=\text{Cay}(\mathcal{z})$ to the original equation $ S(a) = Z J_d  - J_d Z^\top$ can be obtained after solving this vector equation for $\mathcal{z}$ by using (typically 2 or 3 steps of) Newton's method:
\begin{equation}
    \mathcal{z}^{(n+1)} = \mathcal{z}^{(n)}  - \left[ \nabla \phi (\mathcal{z}^{(n)}) \right]^{-1} \phi (\mathcal{z}^{(n)}), \qquad \nabla \phi(\mathcal{z}) = S(a) + (a^\top \mathcal{z}) \mathbb{I}_3 + \mathcal{z}a^\top - 2J.
\end{equation}

\end{remark}

\vspace{-2mm}

 \subsection{Lie Group Forced Variational Integrator Networks (LieFVINs) on \text{SE}(3)} \label{sec: Lie Group Variational Integrator Networks}
 
We now describe the construction of Lie group Forced Variational Integrator Networks (\textcolor{blue}{\textbf{LieFVINs}}), for the forced variational integrator on $\text{SE}(3)$ presented in Section~\ref{sec: Variational Integrator}. The idea is to parametrize the updates of the integrator and match them with observed updates. Here, we consider the case where position-velocity data is available, in which case the LieFVIN is based on equations~\eqref{eq: SE3 discrete H equation 2}-\eqref{eq: SE3 discrete H equation 4}. The case where only position data is available is presented in Appendix~\ref{appendix: position only}. 

We parametrize \textcolor{mygreen}{$m$}, $\mathcal{f}_d^{\pm}$ and $U$ as neural networks. The inertia $J$ is a symmetric positive-definite matrix-valued function of $(x,R)$ constructed via a Cholesky decomposition $J = LL^\top $ for a lower-triangular matrix $L$ implemented as a neural network. Given $J$, we also obtain $J_d = \frac{1}{2} \text{tr}(J) \mathbb{I}_3 - J$. To deal with the implicit nature of equation~\eqref{eq: SE3 discrete H equation 2}, we propose two algorithms, based either on an explicit iterative solver or by penalizing deviations away from equation~\eqref{eq: SE3 discrete H equation 2}: \vspace{-2mm}

\noindent \hrulefill

\noindent \textbf{Algorithm Ia.} Given position-velocity data $\{ (\textcolor{mygreen}{x_0}, R_0, \textcolor{mygreen}{v_0},\omega_0, u_0) \mapsto (\textcolor{mygreen}{x_1},  R_1 , \textcolor{mygreen}{v_1}, \omega_1)  \}$, minimize discrepancies between the observed $ (\textcolor{mygreen}{x_1},  R_1, \textcolor{mygreen}{v_1}, \omega_1)$ quadruples and the predicted $(\textcolor{mygreen}{\tilde{x}_1},\tilde{R}_1, \textcolor{mygreen}{\tilde{v}_1}, \tilde{\omega}_1)$ quadruples, obtained as follows: for each $(\textcolor{mygreen}{x_0}, R_0, \textcolor{mygreen}{v_0}, \omega_0, u_0)$ data tuple,
\begin{enumerate}
\item Get $\mathcal{f}^{R \pm}_{d_0}$ \textcolor{mygreen}{and $\mathcal{f}^{x \pm}_{d_0}$} from $(\textcolor{mygreen}{x_0},R_0,u_0)$, and $ \xi_0$ from $ S(\xi_0) = \frac{\partial U_{0}}{\partial R_{0}}^\top R_{0} - R_{0}^\top \frac{\partial U_{0}}{\partial R_{0}}$
    \item Get $Z_0 = \text{Cay}(\mathcal{z})$ where $\mathcal{z}$ is obtained using a few steps of Newton's method to solve the vector equation~\eqref{eq: Vector Equation General Form} equivalent to 
$  h S(J\omega_0)  +  h S( \mathcal{f}_{d_0}^{R-}) + (1-\alpha) h^2 S(\xi_0) = Z J_d - J_d Z^\top$
\item Compute $\tilde{R}_1 = R_0 Z_0$, and then get $\xi_1$ from $ S(\xi_1) = \frac{\partial U_{1}}{\partial \tilde{R}_{1}}^\top \tilde{R}_{1} - \tilde{R}_{1}^\top \frac{\partial U_{1}}{\partial \tilde{R}_{1}}$
\item Get $\tilde{\omega}_1$ from $ J \tilde{\omega}_1 =  Z_0^\top J\omega_0 + (1-\alpha) h Z_0^\top \xi_0 + \alpha h \xi_{1} + Z_0^\top \mathcal{f}_{d_0}^{R-}+ \mathcal{f}_{d_0}^{R+}  $  
\textcolor{mygreen}{ \item  Compute $\begin{bmatrix}  \tilde{x}_1 \\ \tilde{v}_1  \end{bmatrix} = \begin{bmatrix}  x_0 \\ v_0  \end{bmatrix} + \frac{1}{m} \begin{bmatrix}   h m v_0 - (1-\alpha)h^2 m \frac{\partial U_0}{\partial x_0} -   h R_0 \mathcal{f}_{d_0}^{x-} \\  - (1-\alpha) h^2 \frac{\partial U_0}{\partial x_0} - \alpha h^2 \frac{\partial U_1}{\partial x_1}  + R_0 \mathcal{f}_{d_0}^{x-} +  R_1 \mathcal{f}_{d_0}^{x+}  \end{bmatrix}$} \vspace{-0.7mm}
\end{enumerate} \noindent \hrulefill

\noindent \textbf{Algorithm Ib.}  Given position-velocity data $\{ (\textcolor{mygreen}{x_0},  R_0, \textcolor{mygreen}{v_0}, \omega_0, u_0) \mapsto (\textcolor{mygreen}{x_1},  R_1 , \textcolor{mygreen}{v_1}, \omega_1)  \}$, minimize
\begin{itemize}
    \item Discrepancies between the observed $ (\textcolor{mygreen}{x_1}, \textcolor{mygreen}{v_1}, \omega_1)$ triples and the predicted $(\textcolor{mygreen}{\tilde{x}_1}, \textcolor{mygreen}{\tilde{v}_1}, \tilde{\omega}_1)$ triples
    \item Deviations away from the equation $  h S(J\omega_0)  +  h S( \mathcal{f}_{d_0}^{R-}) + (1-\alpha)h^2 S(\xi_0) = J_d Z_0 - Z_0^\top J_d$
\end{itemize}
where, for each $(\textcolor{mygreen}{x_0},  R_0, \textcolor{mygreen}{v_0}, \omega_0, u_0, R_1)$ data tuple,
\begin{enumerate}
     \item $\mathcal{f}^{R \pm}_{d_0}$ \textcolor{mygreen}{and $\mathcal{f}^{x \pm}_{d_0}$} are obtained from $(\textcolor{mygreen}{x_0},R_0,u_0)$, and $ \xi_0,\xi_1$ from $ S(\xi_k) = \frac{\partial U_{k}}{\partial R_{k}}^\top R_{k} - R_{k}^\top \frac{\partial U_{k}}{\partial R_{k}}$
     \item $Z_0 = R_0^\top R_1$ and $\tilde{\omega}_1 = J^{-1}\left[  Z_0^\top J\omega_0 + (1-\alpha)h Z_0^\top \xi_0 + \alpha h \xi_{1} + Z_0^\top \mathcal{f}_{d_0}^{R-}+ \mathcal{f}_{d_0}^{R+} \right] $      \vspace{-0.7mm}       \textcolor{mygreen}{\item  $\begin{bmatrix}  \tilde{x}_1 \\ \tilde{v}_1  \end{bmatrix} = \begin{bmatrix}  x_0 \\ v_0  \end{bmatrix} + \frac{1}{m} \begin{bmatrix}   h m v_0 - (1-\alpha)h^2 m \frac{\partial U_0}{\partial x_0} -   h R_0 \mathcal{f}_{d_0}^{x-} \\  - (1-\alpha) h^2 \frac{\partial U_0}{\partial x_0} - \alpha h^2 \frac{\partial U_1}{\partial x_1}  + R_0 \mathcal{f}_{d_0}^{x-} +  R_1 \mathcal{f}_{d_0}^{x+}  \end{bmatrix}$}  \vspace{-1mm}
\end{enumerate} \noindent \hrulefill

This general strategy extends to any other Lie group integrator. In particular, LieFVINs on $\text{SO}(3)$ can be obtained from the algorithms above as the special case where $x$ is constant, in which case we can disregard all the variables and operations in \textcolor{mygreen}{green}. Lie group variational integrator networks without forces (\textcolor{blue}{\textbf{LieVINs}}) can be obtained by setting $\mathcal{f}_{d_0}^{R\pm} = \mathcal{f}_{d_0}^{x\pm} = 0$. Note that the strategy behind Algorithm Ia enforces the structure of the system in a stronger way than in Algorithm Ib. However, for certain Lie groups and variational integrators, it might not be practical to use Newton's method to solve for the implicit updates, in which case Algorithm Ib is preferred.

\subsection{Control Strategy}
\label{subsec:control_strategy}


Given the discrete-time flow map $\Psi$ learnt by a LieFVIN, we can formulate a Model Predictive Control (MPC) problem to design a discrete-time control policy for the dynamical system:  \vspace{-2mm}

\noindent \hrulefill

\noindent At each step $t_\ell  = \ell h$,
\begin{enumerate}
    \item Obtain an estimate $(\tilde{q}_\ell , \dot{\tilde{q}}_\ell)$ of the current state.
    
    \item Solve a $N$-step finite horizon optimal control problem starting at $(\tilde{q}_\ell , \dot{\tilde{q}}_\ell)$, formulated as a constrained optimization problem: \vspace{-2mm} \textit{Minimize the discrete cost function} 
\begin{equation} \label{eq:mpc problem} \mathcal{J}_d  (U_{\ell}) = \sum_{k =0}^{N-1}{\mathcal{C}_d(q_{\ell+k},q_{\ell+k+1},\dot{q}_{\ell+k},u_{\ell +k})} + \Phi_d(q_{\ell + N-1},q_{\ell +N},\dot{q}_{\ell + N},u_{\ell +N-1} ), \vspace{-1.6mm}
\end{equation}
\textit{over admissible discrete controls} $U_{\ell} = \{u_\ell, u_{\ell +1} , ..., u_{\ell + N-1}  \}$, \textit{subject to path constraints} $
    \mathcal{P}_d(q_{\ell +k},q_{\ell + k+1},\dot{q}_{\ell +k},u_{\ell +k}) \geq  0$  \textit{for }   $k=1,...,N-1$ \textit{and to the termination condition} $\mathcal{T}_d(q_{\ell + N-1},q_{\ell +N},\dot{q}_{\ell +N},u_{\ell + N-1}) = 0,$
\textit{and where the evolution of the controlled system is prescribed by the surrogate symplectic map $\Psi$ learnt by the LieFVIN.}  
    
    \item Apply the resulting optimal control $u^*_\ell $ to the system in state $(\tilde{q}_\ell , \dot{\tilde{q}}_\ell)$ until $t_{\ell +1} = (\ell +1)h$.
\end{enumerate}  \vspace{-5.2mm}
\noindent \hrulefill

Note that the Lie group constraints do not need to be added as path constraints since they are automatically satisfied to (almost) machine precision, by the design of the LieFVINs. 
In our experiments, we use the PyTorch MPC framework\footnote{Code: \url{https://locuslab.github.io/mpc.pytorch/}} \citep{tassa2014control, amos2018differentiable}.

\section{Evaluation}
\label{sec:exp_results}

We now demonstrate our approach to learn and control a planar pendulum and a crazyflie quadrotor. More details about our implementation can be found in Appendix~\ref{appendix: Implementation}, and our Python/PyTorch code is available at    \ \url{https://thaipduong.github.io/LieFVIN/}

\subsection{Pendulum}
\label{subsec:pendulum_so3}
We consider a planar pendulum with dynamics $ \ddot{\varphi} = -15\sin{\varphi} + 3u,$ where $\varphi$ is the angle with respect to its downward position $\varphi = 0$ and $u \in \mathbb{R}$ is a control input. The mass of the pendulum, the potential energy, and input coefficient are given by $m = 1/3$, $U(\varphi) = 5(1-\cos{\varphi})$,  $g(\varphi) = 1$. We collected $\{(\cos{\varphi}, \sin{\varphi}, \dot{\varphi})\}$ data from an OpenAI Gym environment \citep{Zhong2020}. LieFVIN was trained with position-velocity data as described in Algorithm Ia with $\alpha = 0.5$. The forces were specified as $\mathcal{f}_d^{R+} = 0$ and $\mathcal{f}_d^{R-} = g(q)u$, where $g(q)$ is a neural network. 

\begin{figure}[ht!]	
\vspace{-0.5mm}
\centering
	\includegraphics[width=0.245\textwidth]{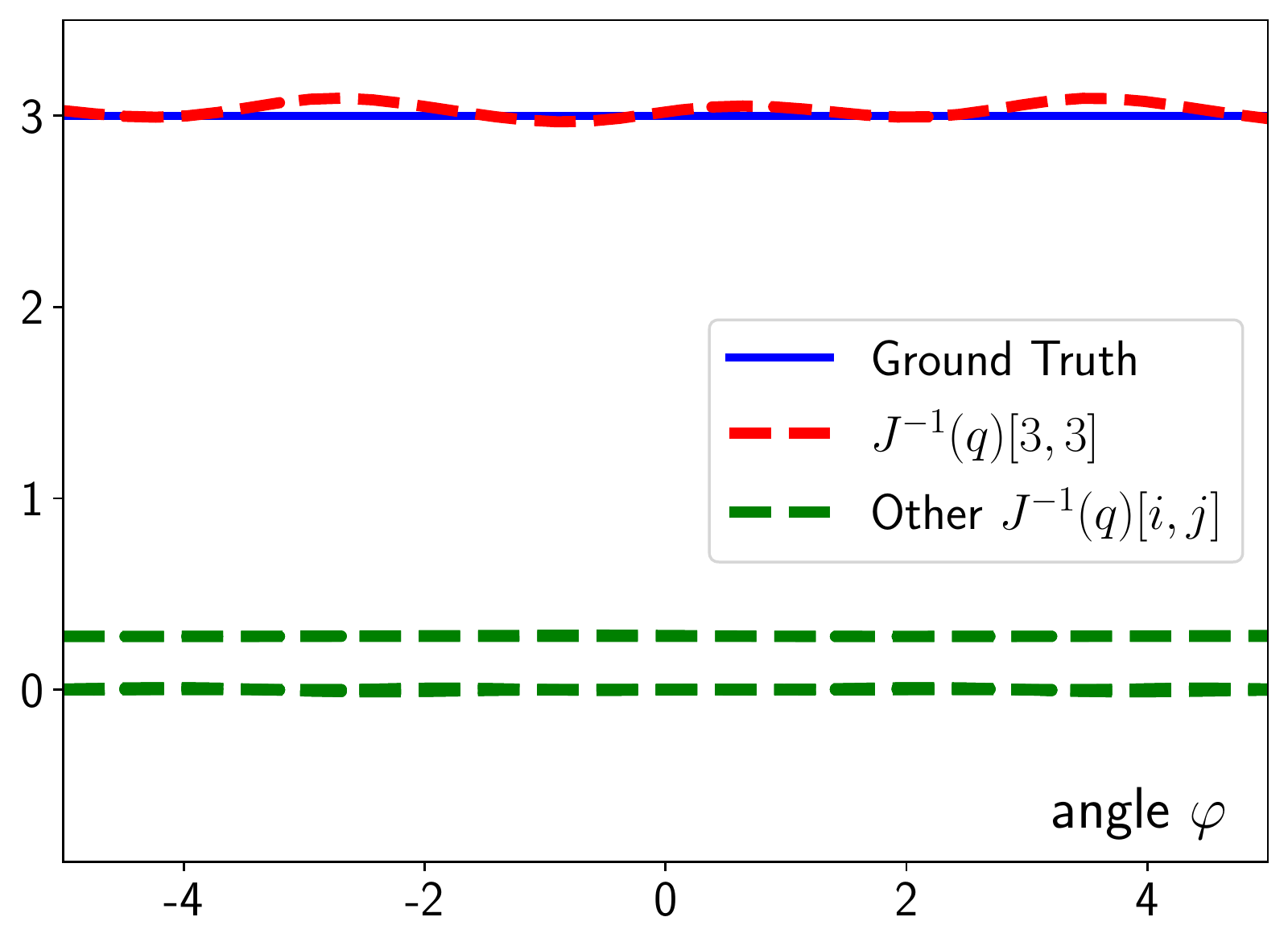}
	\includegraphics[width=0.245\textwidth]{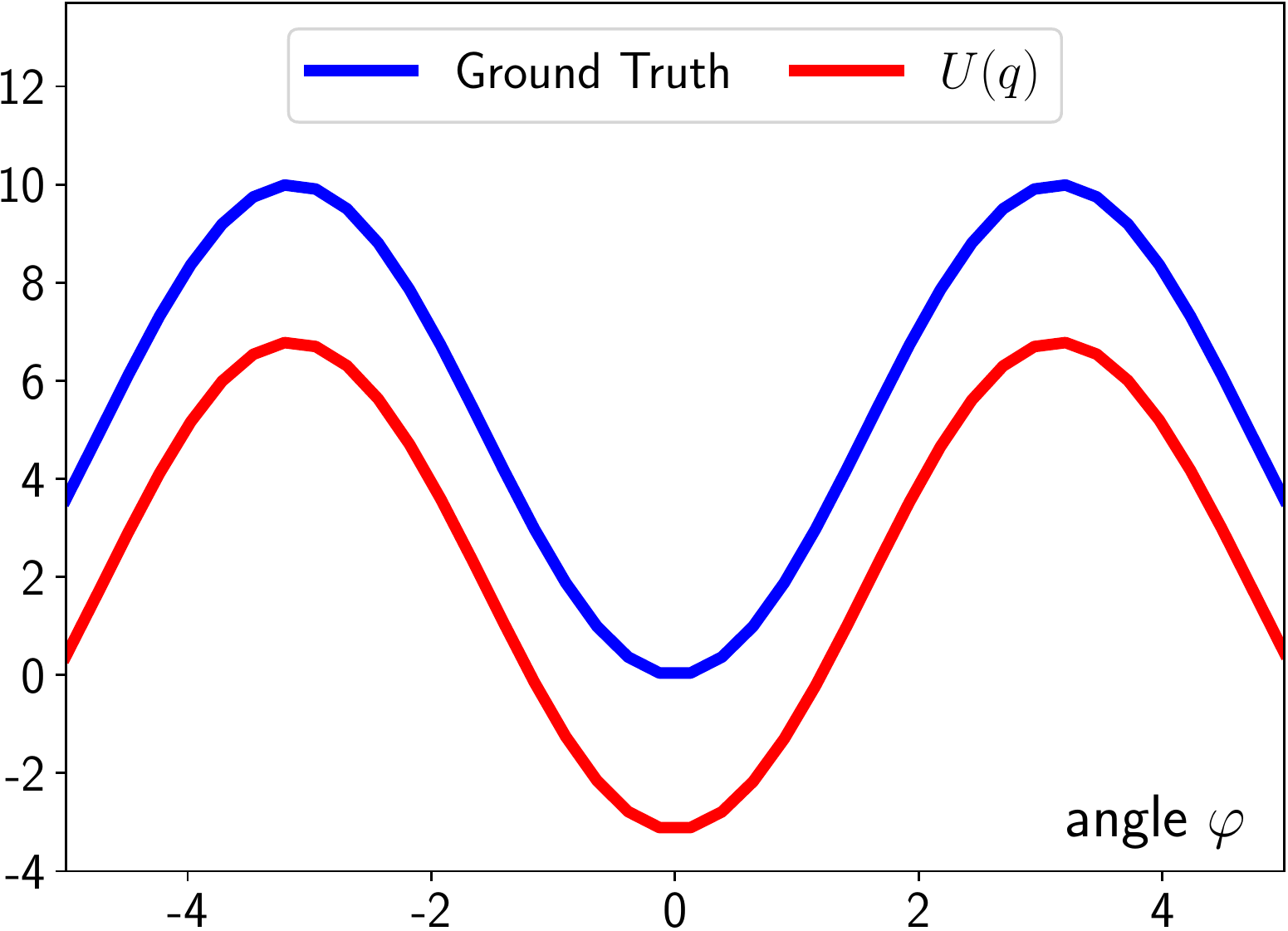}
	\includegraphics[width=0.245\textwidth]{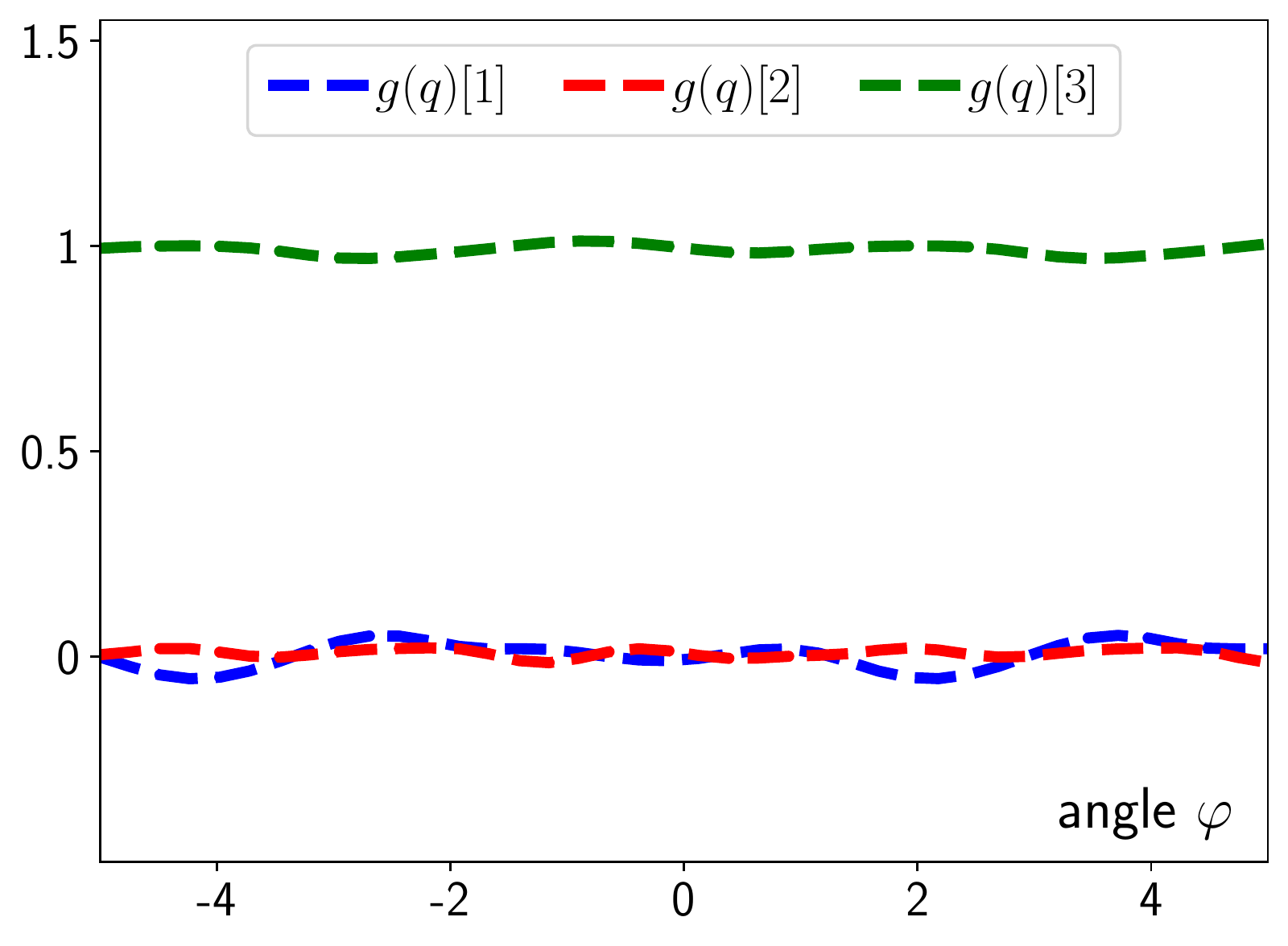}
	\includegraphics[width=0.245\textwidth]{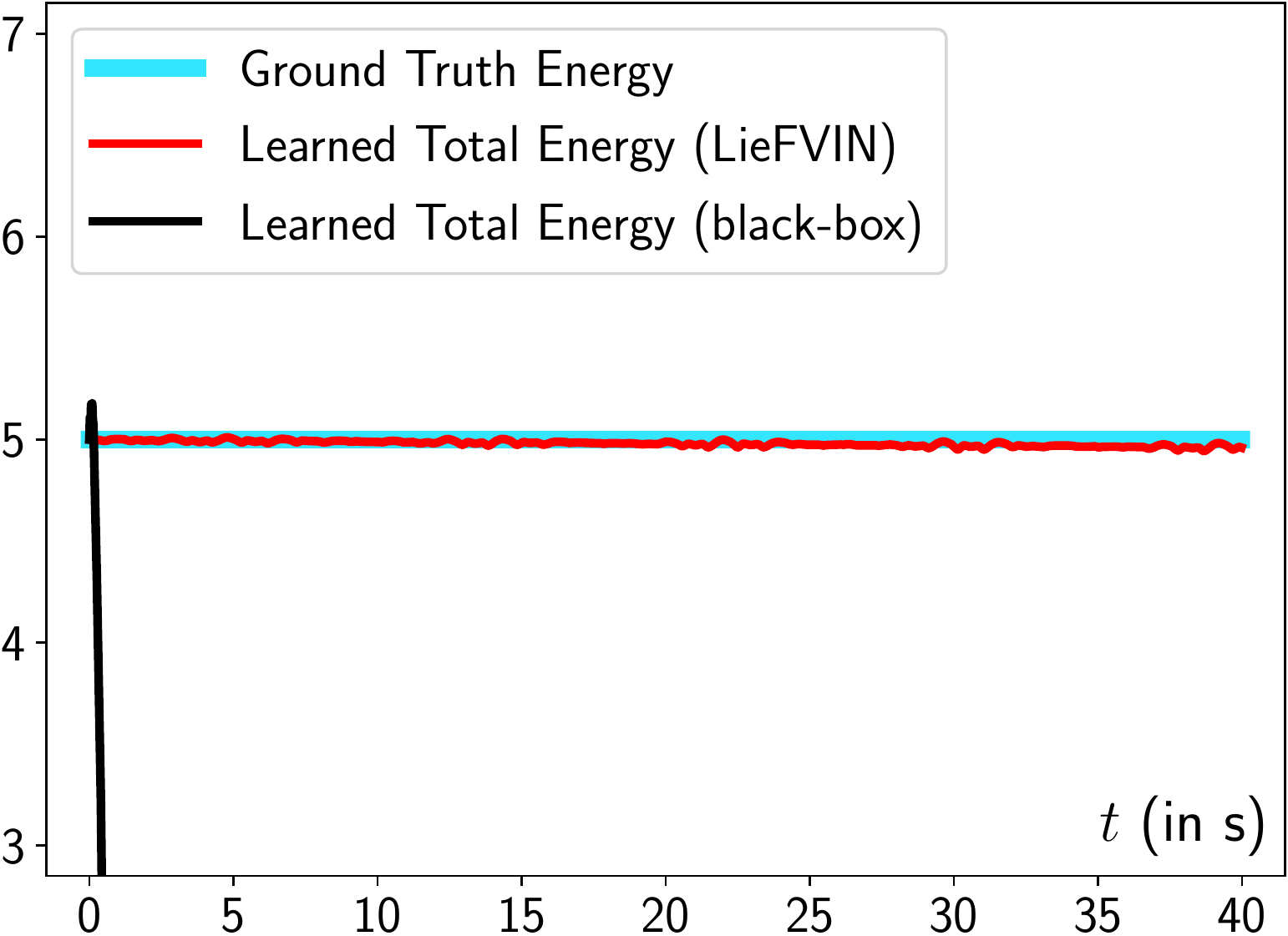}
	\put(-428,9.5){\small \textit{(a)}}	
	\put(-317,9.5){\small \textit{(b)}}
	\put(-206,9.5){\small \textit{(c)}}	
	\put(-95,9.5){\small \textit{(d)}}
			
	\includegraphics[width=0.245\textwidth]{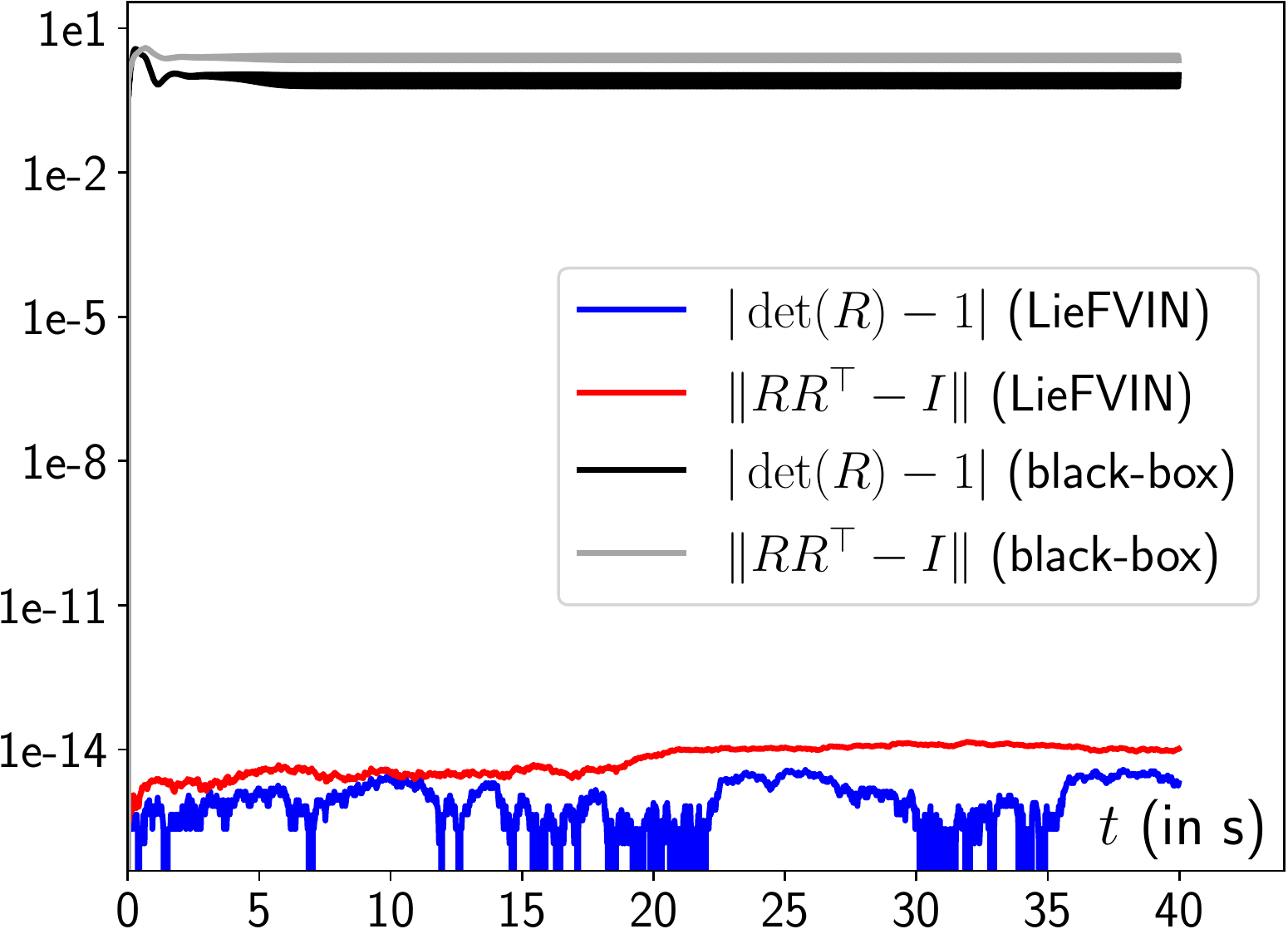}
	\includegraphics[width=0.245\textwidth]{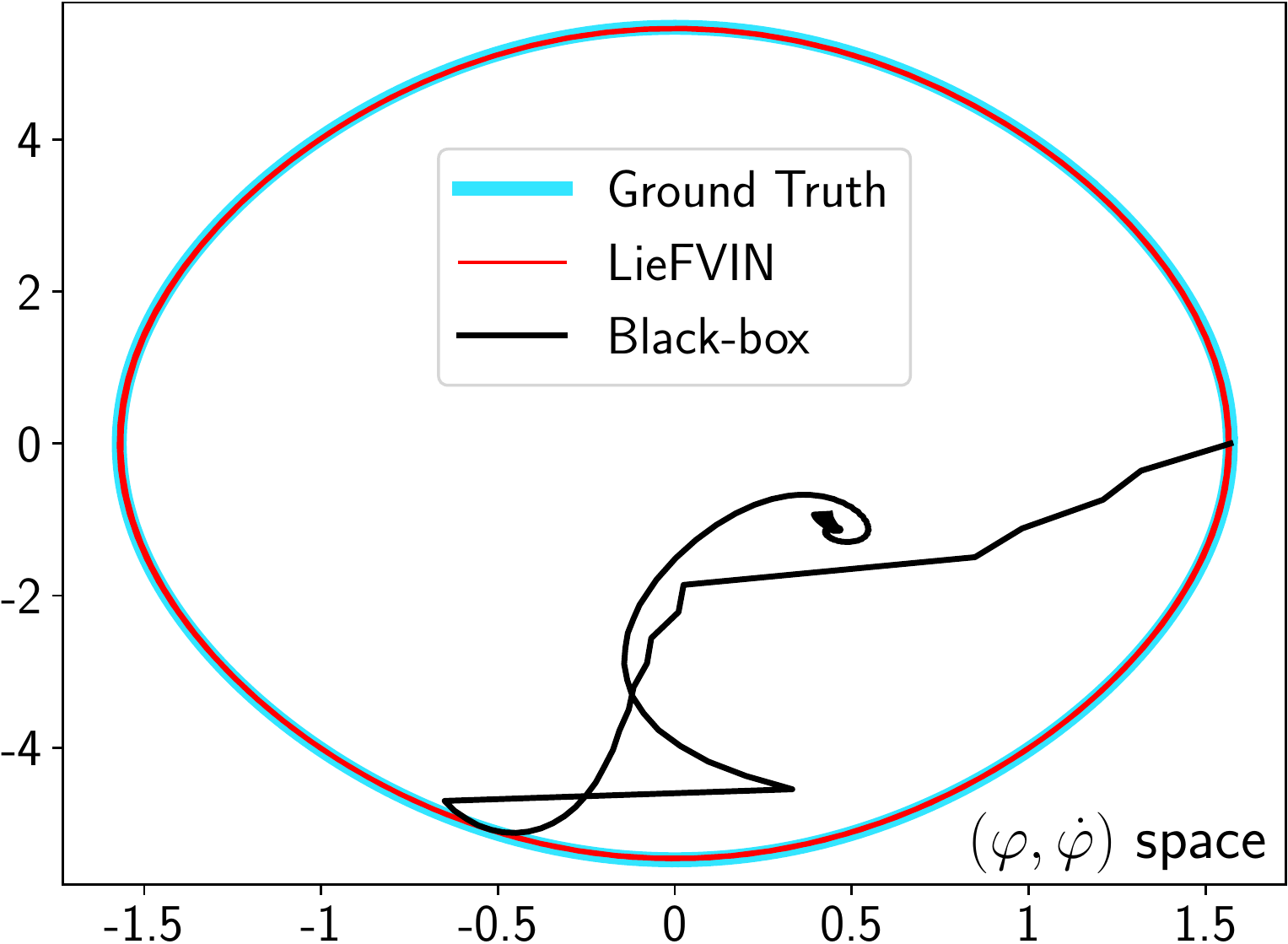}
	\includegraphics[width=0.245\textwidth]{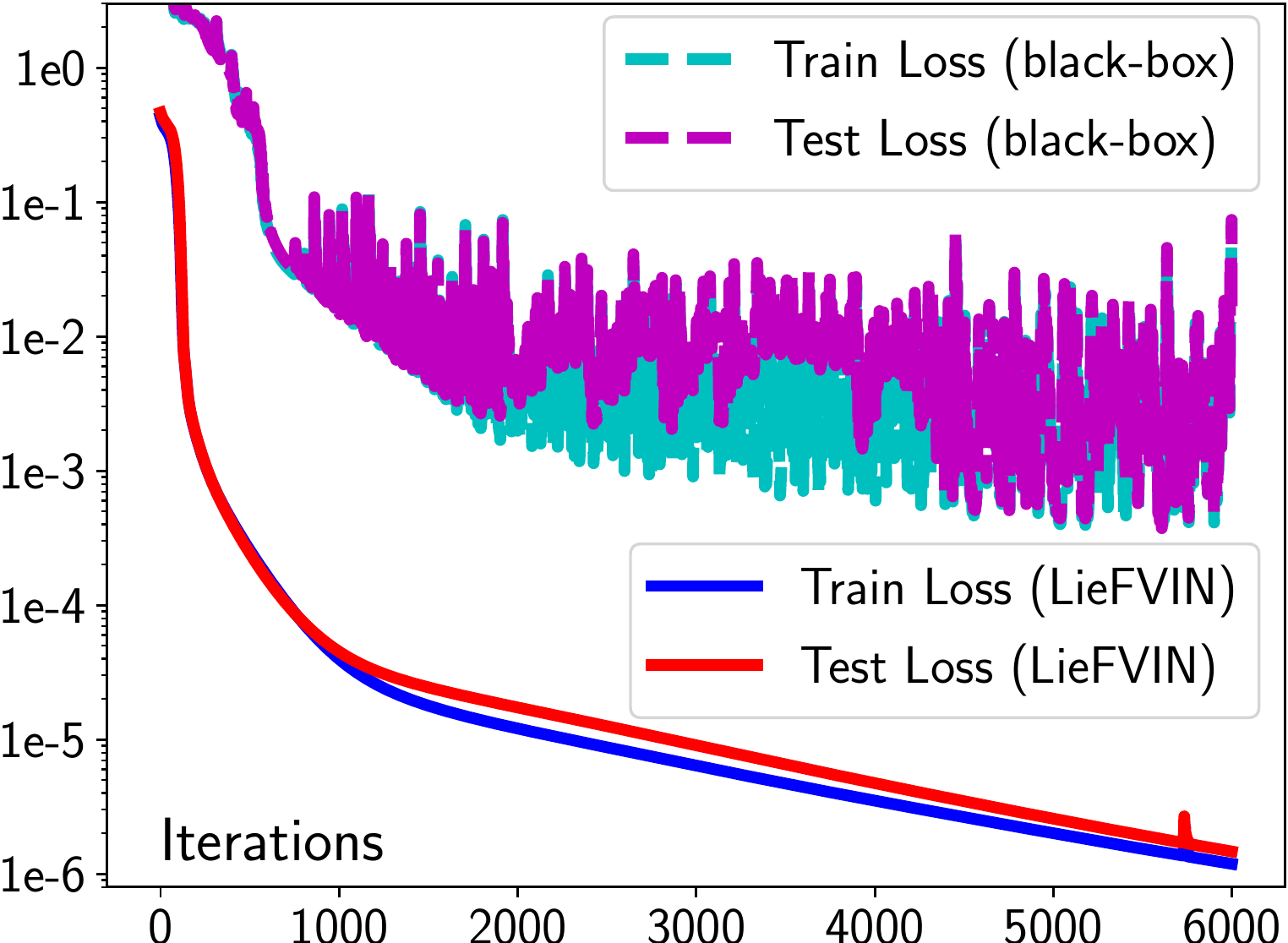}
		\includegraphics[width=0.245\textwidth]{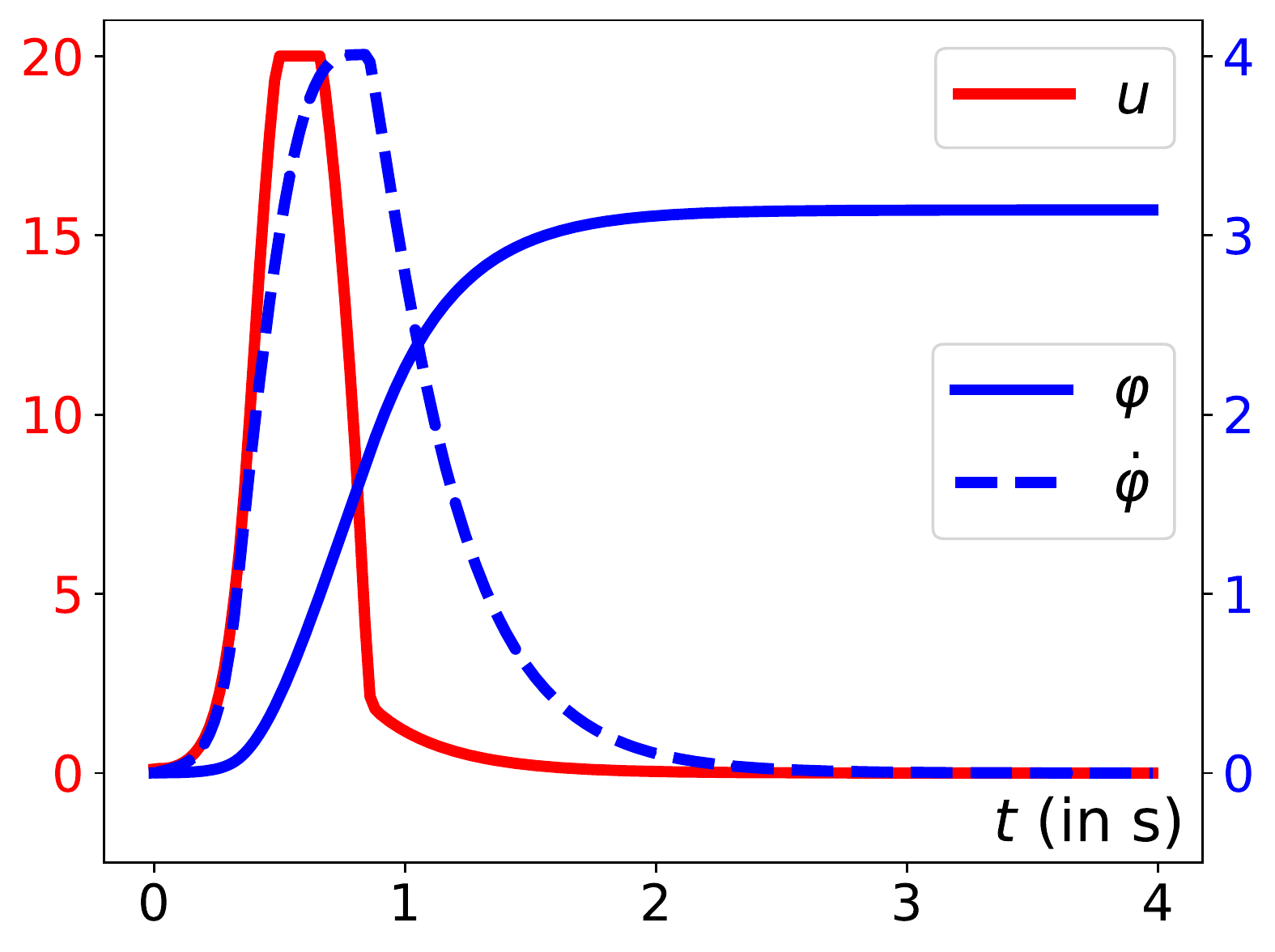}
	\put(-420,58){\small \textit{(e)}}	
	\put(-315.5,67){\small \textit{(f)}}
	\put(-190,67){\small \textit{(g)}}	
	\put(-97.6,67){\small \textit{(h)}}

\vspace*{-3mm}	
\caption{Evaluation of $\text{SO}(3)$ LieFVIN on a pendulum. We learned the inertia matrix~(a), potential energy~(b), and input coefficient~(c), with the loss function shown in~(g). The learnt model respects the energy conservation law~(d), $\text{SO}(3)$ constraints~(e), and phase portraits~(f). The control from MPC is shown in (h). Meanwhile, a black-box model struggles to infer the $\text{SO}(3)$ constraints from data (e)(g) and is not able to conserve energy (d). }
\label{fig:pend_exp} \vspace*{-1mm}	
\end{figure}

Figures \ref{fig:pend_exp}(a), (b), (c) show that the LieFVIN model learned the correct inertia matrix $J$, control gain $g(q)$, and potential energy $U$ (up to a constant offset). Without control input, i.e., $\mathcal{f}_d^{R\pm} = 0$, we use the dynamics model learnt from short-term trajectories of $10$ steps of $0.02$s to generate long-term predictions ($2000$ steps, i.e. $40$s). Figure~\ref{fig:pend_exp}(d) shows that the total energy of the learnt system fluctuates but stays close to the ground truth value. The fluctuation comes from the discretization errors in equations \eqref{eq: SE3 discrete H equation 2}-\eqref{eq: SE3 discrete H equation 4} and model errors for the learnt quantities $J$, $U$, and $g(q)$. Note that the $\text{SO}(3)$ constraint errors remain very small, around $10^{-14}$ (see Figure~\ref{fig:pend_exp}(e)). The phase portraits and the learnt dynamics are close to the ground-truth ones, illustrating the ability to generate long-term predictions using the model learnt from short-term data. Meanwhile, a Multilayer Perceptron black-box model, described in Appendix~\ref{appendix: Implementation Pendulum}, struggles to infer the $\text{SO}(3)$ constraints from data (see Figure~\ref{fig:pend_exp}(e)(g)) and is not able to conserve the total~energy~(see Figure~\ref{fig:pend_exp}(d)).

The learnt dynamics model is combined with MPC as described in Section \ref{subsec:control_strategy} to drive the pendulum from downward position $\varphi = 0$ to a stabilized upright position $\varphi^* = \pi$, $\dot{\varphi}^* = 0$, with input constraint $|u| \leq 20$. Figure~\ref{fig:pend_exp}(h) plots the angle $\varphi$, angular velocity $\dot{\varphi}$, and control input $u$, showing that the pendulum is successfully stabilized using the learnt discrete dynamics model.

\subsection{Crazyflie Quadrotor}

We demonstrate that our $\text{SE}(3)$ dynamics learning and control approach can achieve trajectory tracking for an under-actuated system by considering a Crazyflie quadrotor simulated using PyBullet \citep{gym-pybullet-drones2020}. The control input $\bfu = [f, \bftau]$ includes the thrust $f\in \mathbb{R}_{\geq 0}$ and torque vector $\bftau \in \mathbb{R}^3$ generated by the $4$ rotors. LieFVIN is trained as in Algorithm Ib with $\alpha = 0.5$. The forces are specified as $\mathcal{f}_d^{x\pm} = 0.5g_x(q)u$ and $\mathcal{f}_d^{R\pm} = 0.5g_R(q)u$ where $g_x(q),  g_R(q)$ are neural networks. 

\begin{figure}[ht!]		
\centering
	\includegraphics[width=0.245\textwidth]{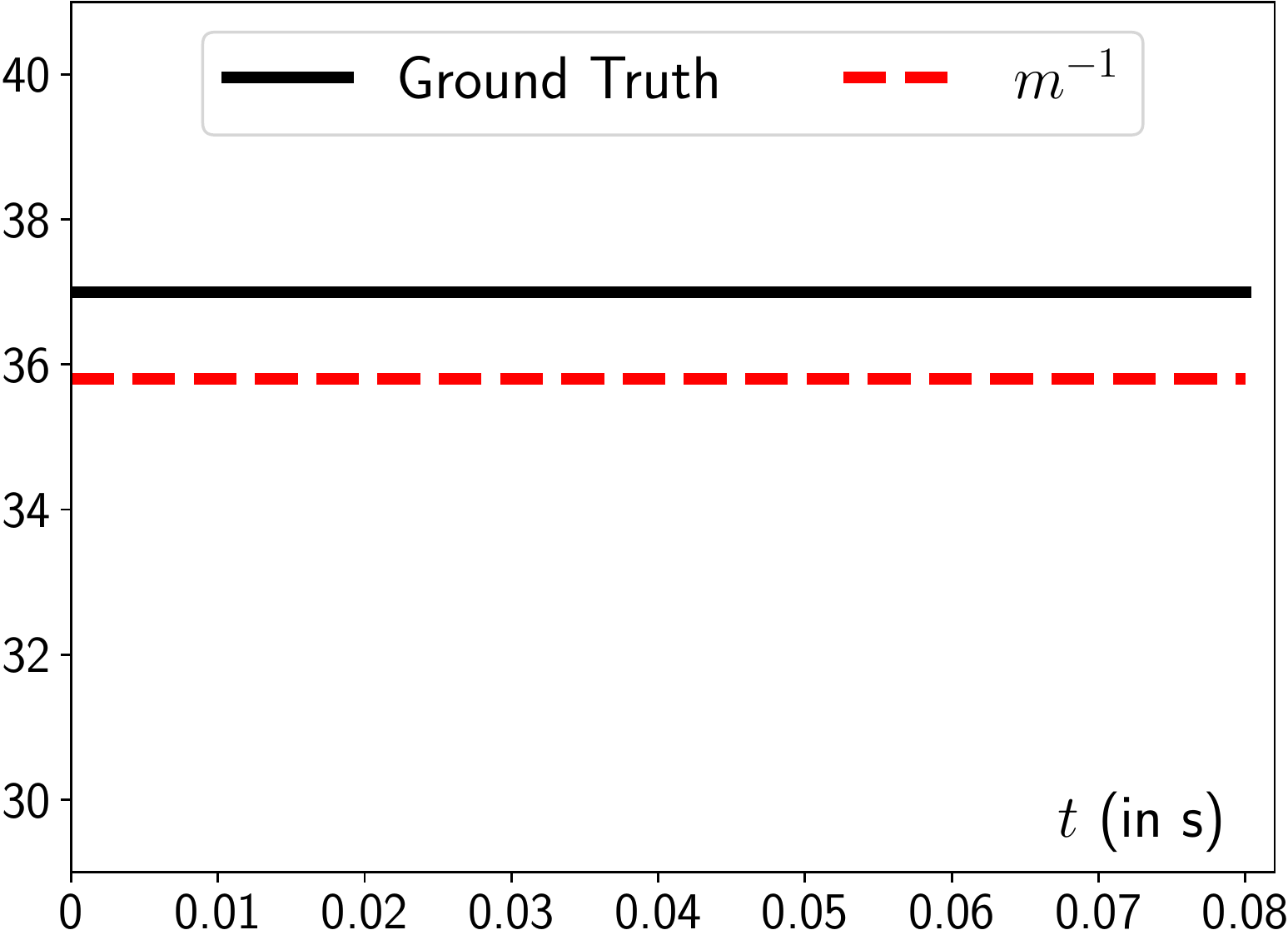}
        \includegraphics[width=0.245\textwidth]{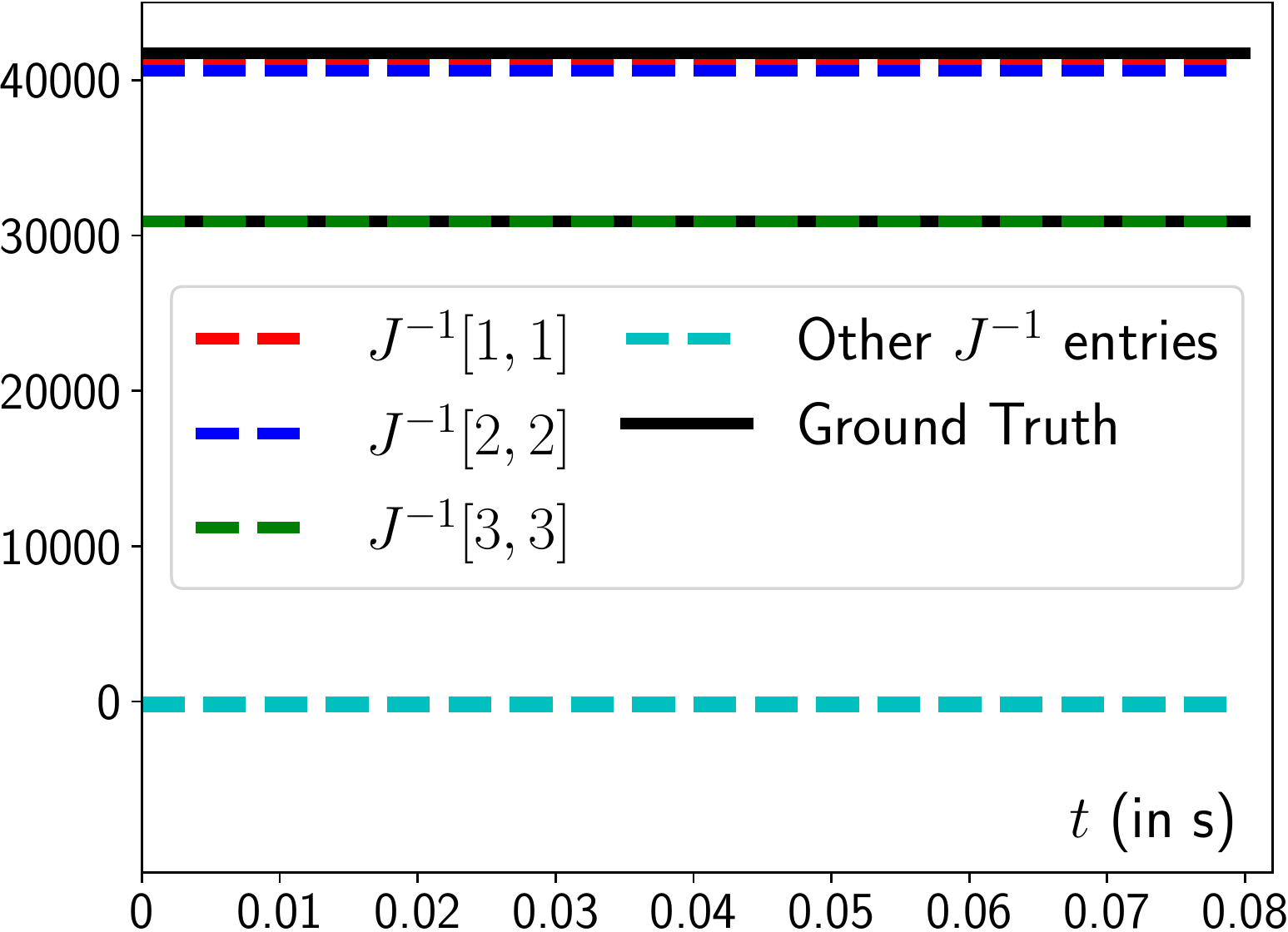}
	\includegraphics[width=0.245\textwidth]{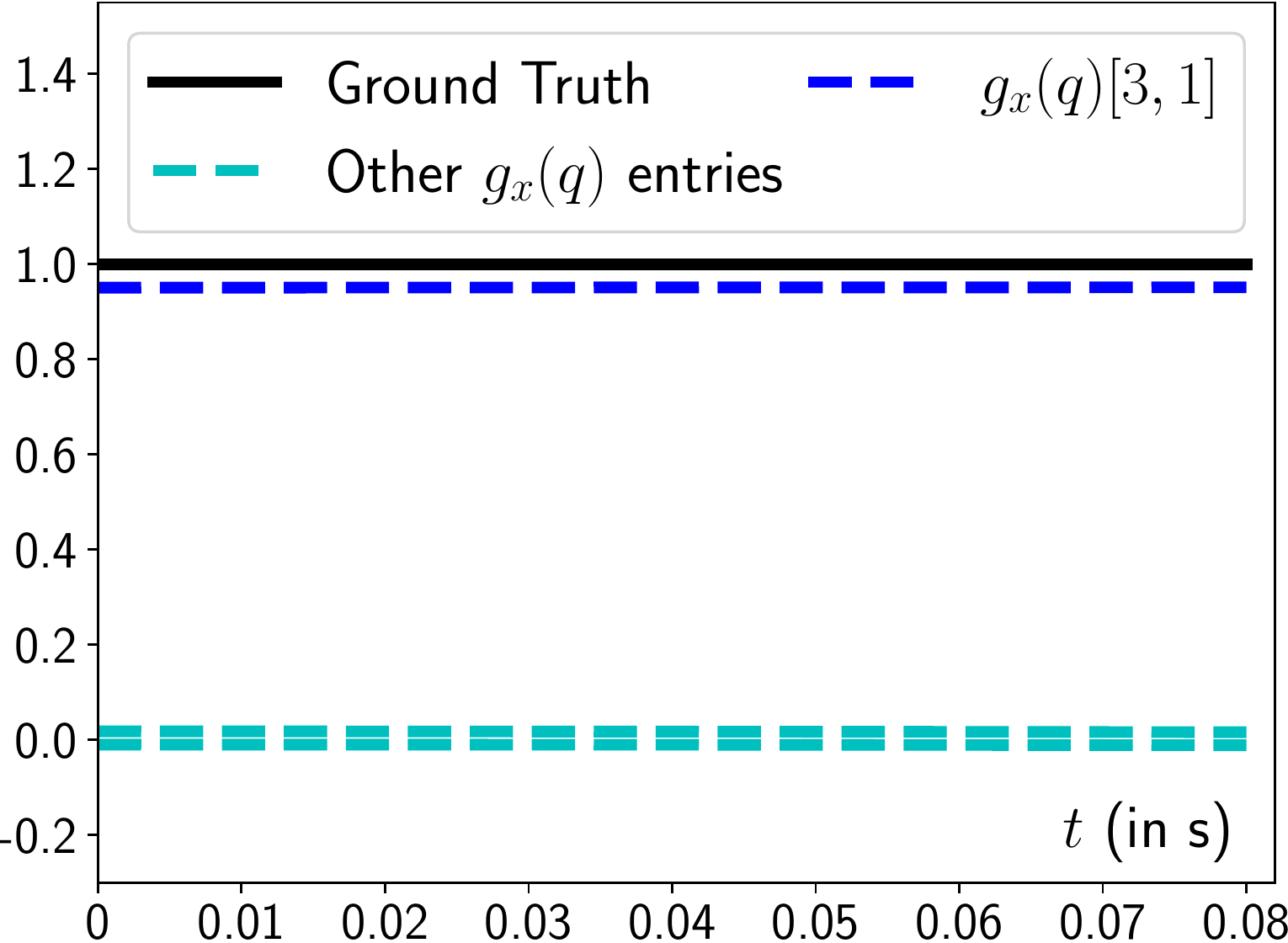}
        \includegraphics[width=0.245\textwidth]{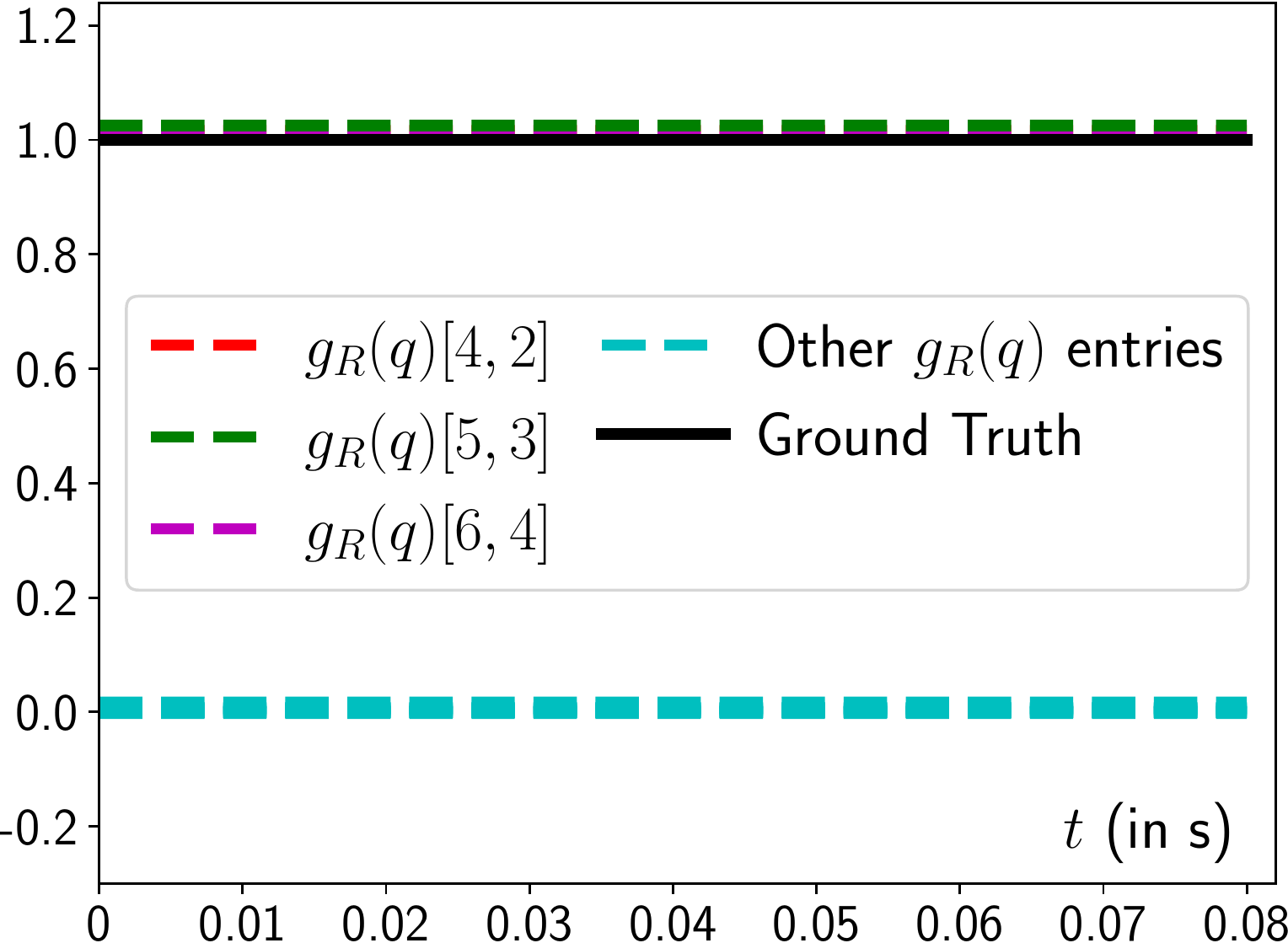}
	\put(-425,8){\small \textit{(a)}}	
	\put(-311,8){\small \textit{(b)}}
	\put(-205,8){\small \textit{(c)}}	
	\put(-96,8){\small \textit{(d)}}

        \includegraphics[width=0.245\textwidth]{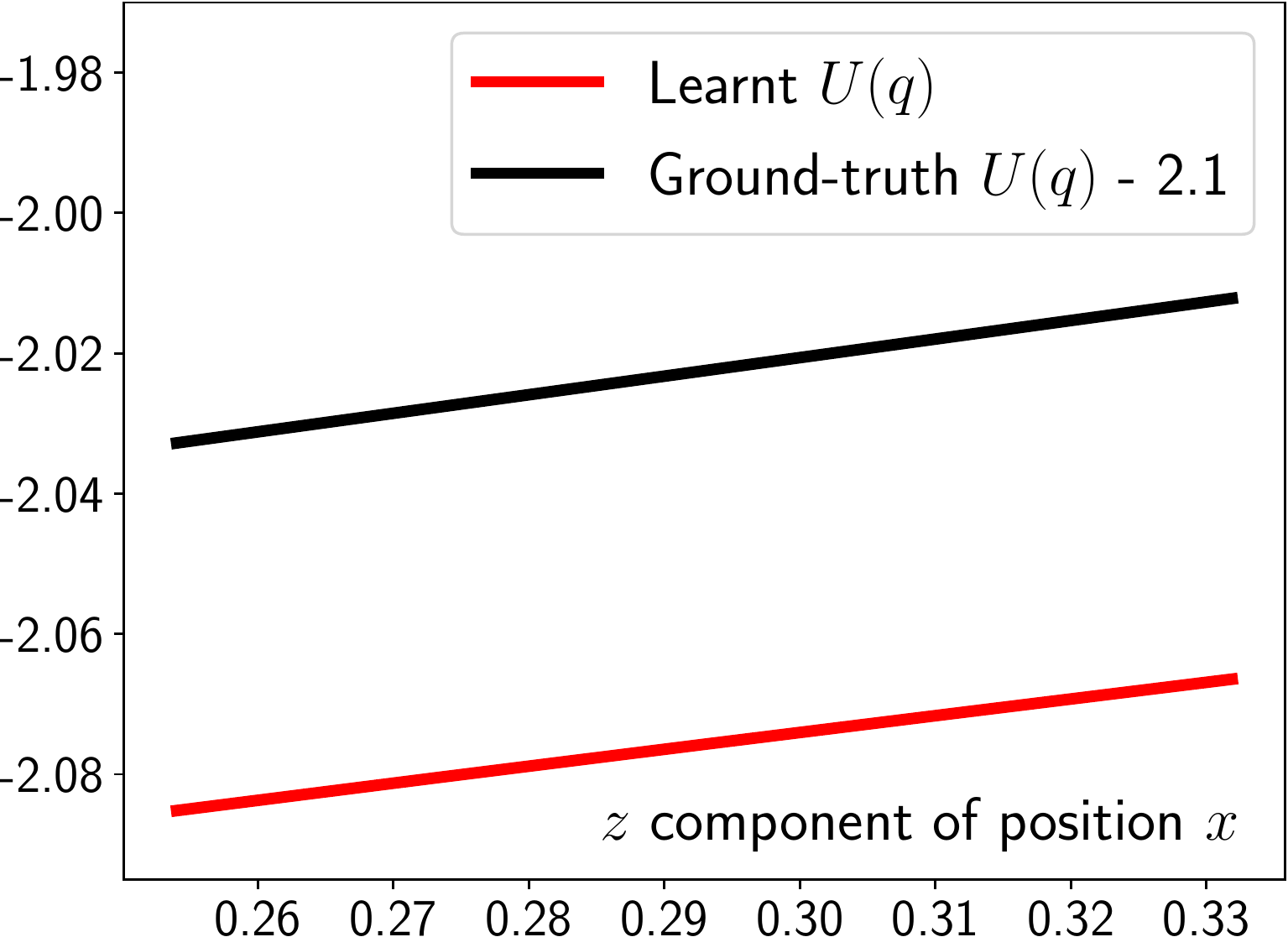}
	\includegraphics[width=0.245\textwidth]{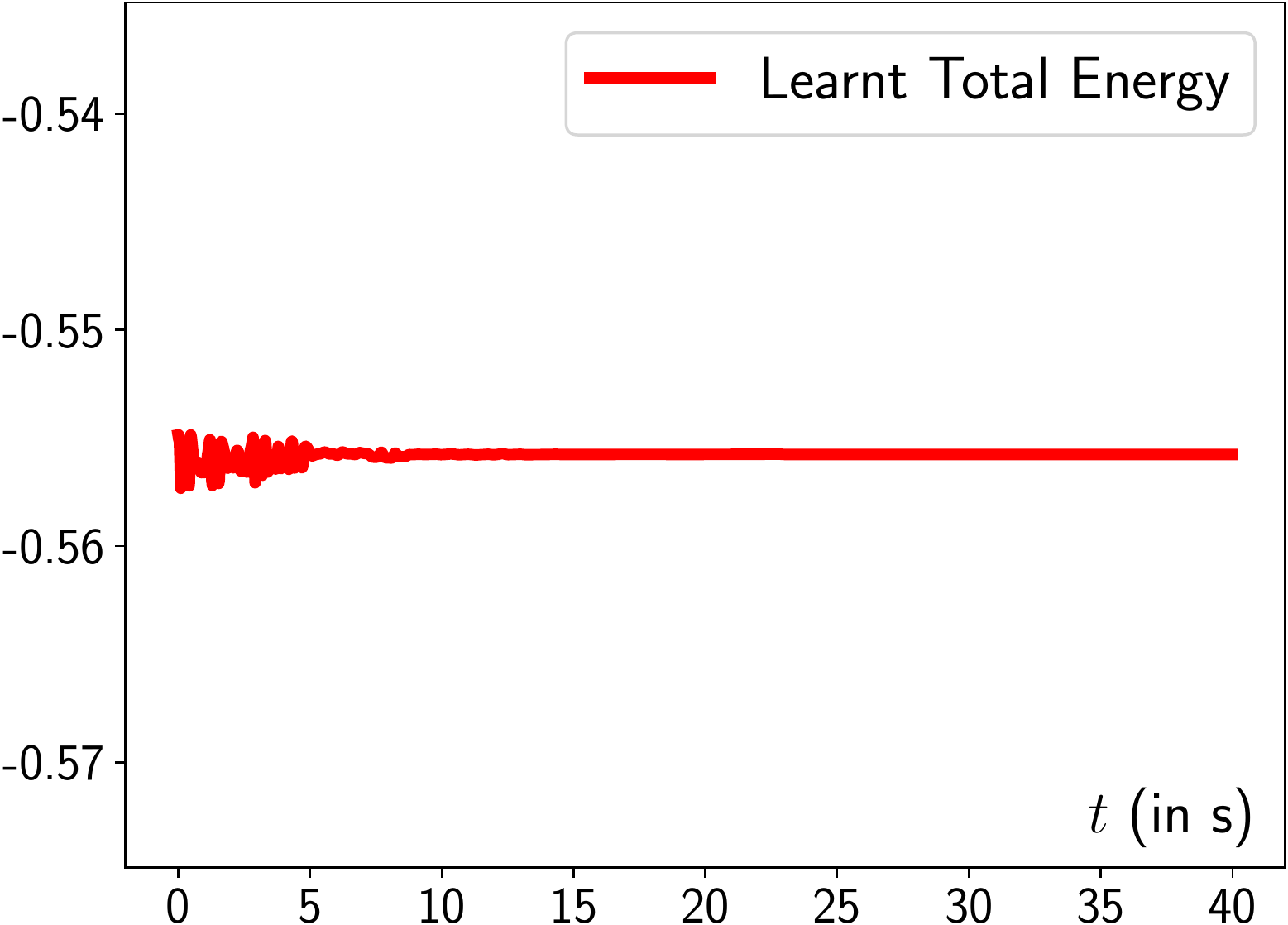}
	\includegraphics[width=0.245\textwidth]{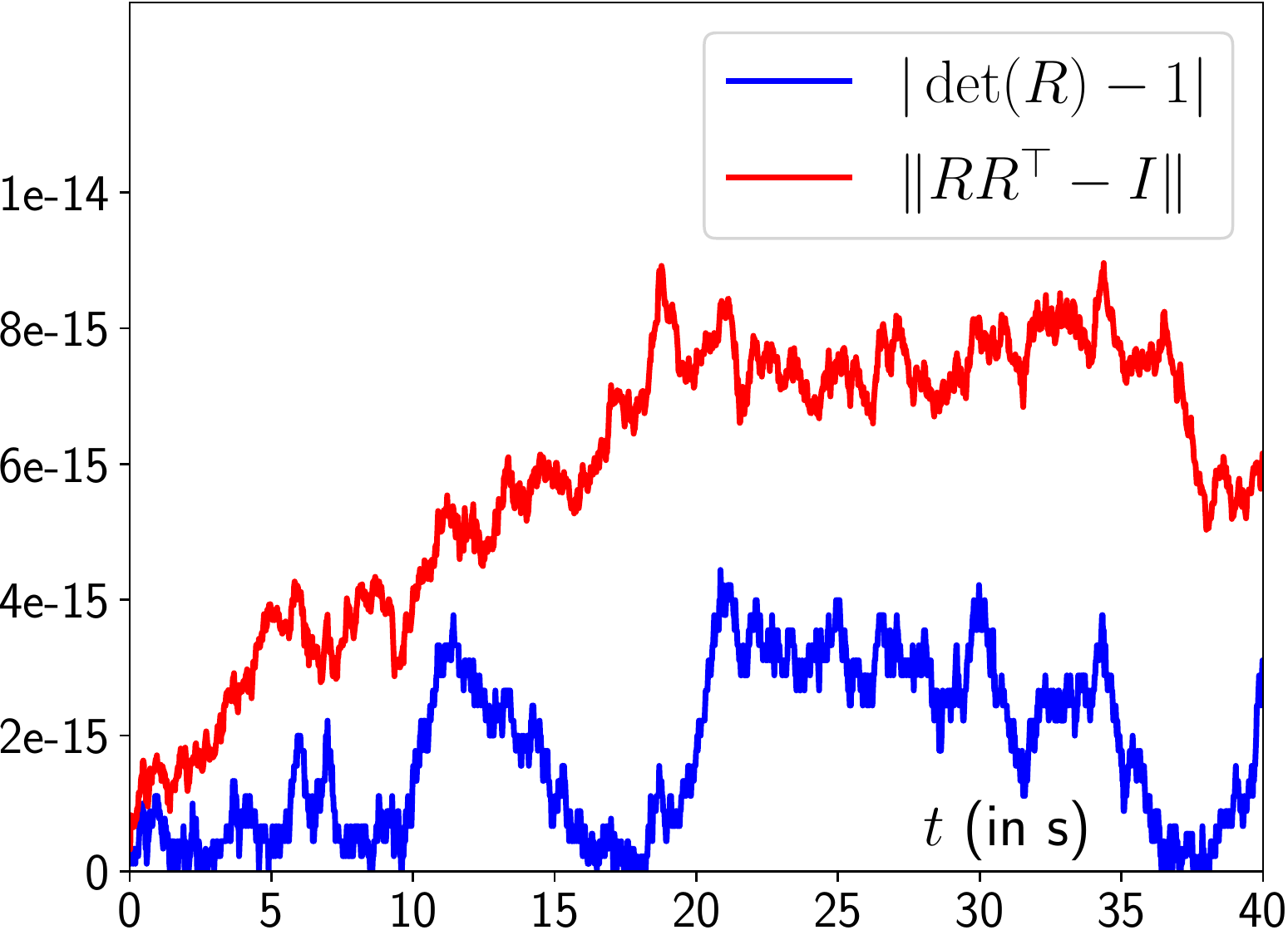}
	\includegraphics[width=0.245\textwidth]{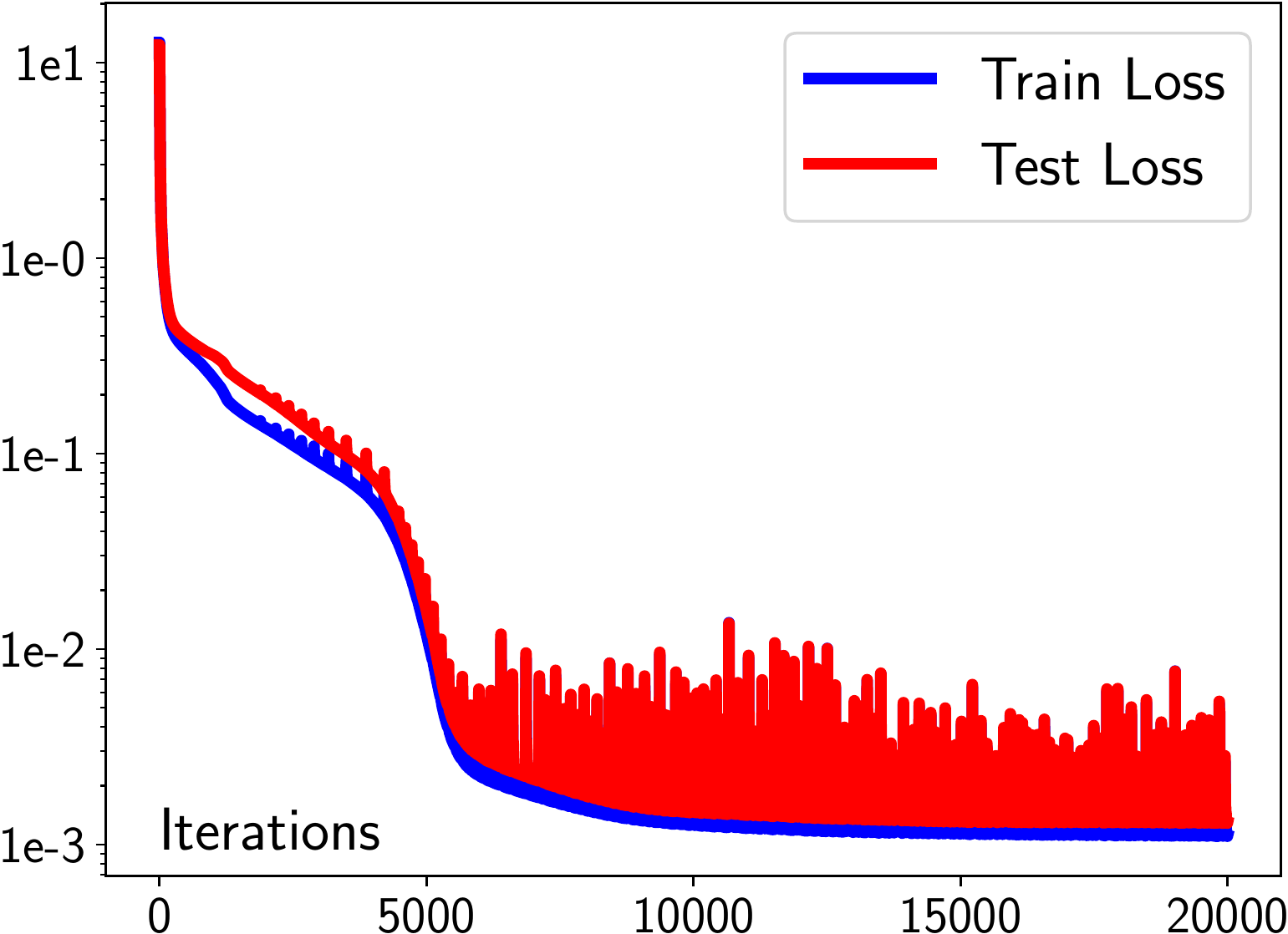}
	\put(-420,67){\small \textit{(e)}}	
	\put(-311,67){\small \textit{(f)}}
	\put(-201.5,67){\small \textit{(g)}}	
	\put(-90,67){\small \textit{(h)}}

\vspace{-4.1mm}	
\caption{LieFVIN learns the correct mass $m$~(a), inertia matrix $J$~(b), input coefficients $g_x(q)$~(c) and $g_R(q)$~(d), potential energy $U(q)$~(e). The learnt model respects the energy conservation law~(f), $\text{SO}(3)$ constraints~(g). The evolution of the loss function is shown in~(h).}
\label{fig:quad_exp}
\end{figure}

Figures \ref{fig:quad_exp}(a)-(e) show that LieFVIN learned the correct mass~$m$, inertia matrix~$J$, control gains~$g_x(q)$ and~$g_R(x)$, and potential energy~$U(q)$ (up to a constant offset). Without control input, i.e., $\mathcal{f}_d^{R\pm} = 0$, we use the dynamics model learnt from short-term trajectories of $5$ steps of $0.02$s to generate long-term predictions ($2000$ steps, i.e. $40$s). Figure~\ref{fig:quad_exp}(f) shows that the total energy of the system has bounded fluctuations while $\text{SO}(3)$ constraint errors are around~$10^{-14}$, verifying the near-energy conservation and manifold constraints guaranteed by our approach.

\begin{figure}[ht!]	
\vspace{-2.5mm}	
\centering
\includegraphics[width=0.73\textwidth]{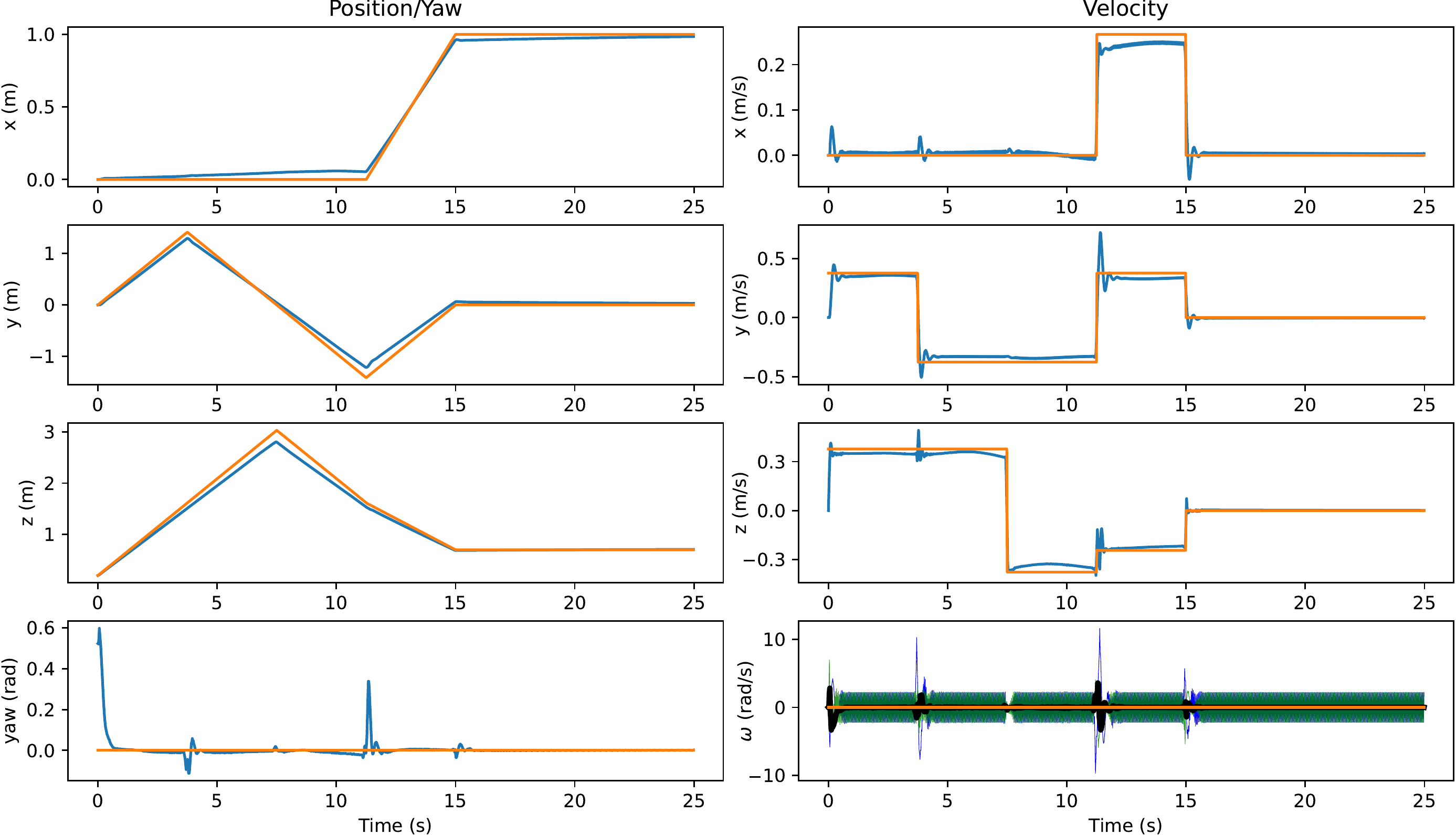}
 \hfill
         \begin{minipage}[h]{0.26\textwidth}
         \vspace{-6.1cm}
         \centering
        \includegraphics[width=0.83\textwidth]{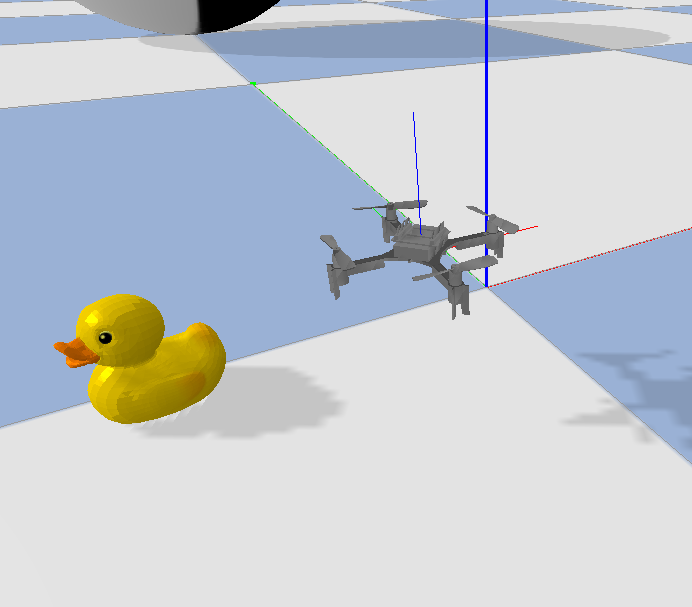}
        \includegraphics[width=0.93\textwidth]{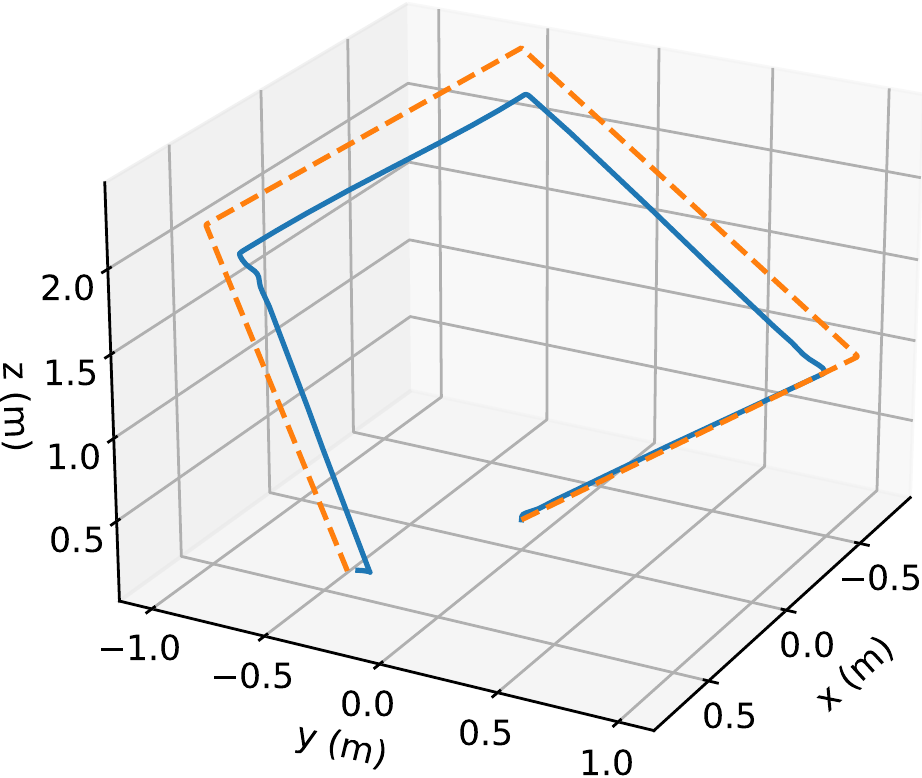}

        \end{minipage}
        
\vspace{-4.2mm}	
\caption{Trajectory tracking with the learned quadrotor model. The tracking errors (left) between reference trajectory (orange) and the actual trajectory, and the robot trajectory (lower right) show that the task is completed successfully.} 
\label{fig:tracking_viz}  
\end{figure}

The learnt model is then combined with MPC as in Section \ref{subsec:control_strategy} to track a diamond-shaped trajectory, with control input constraints $0\leq f \leq 0.595, \  \vert \tau \vert \leq 10^{-3}[5.9 \ \ 5.9 \ \ 7.4 ]^\top$. Figure~\ref{fig:tracking_viz} displays the robot trajectory and plots the tracking errors over time, showing that the quadrotor successfully completes the task.


\newpage

\section{Conclusion}
\label{sec:conclusion}

We introduced a new structure-preserving deep learning strategy to learn discrete-time flow maps for controlled Lagrangian or Hamiltonian dynamics on a Lie group, from position-velocity or position-only data. The resulting maps evolve intrinsically on the Lie group and preserve the symplecticity underlying the systems of interest, which allows to generate physically well-behaved long-term predictions based on short-term trajectories data. Learning discrete-time flow maps instead of vector fields yields better prediction without requiring the use of a numerical integrator, neural ODE, or adjoint techniques. The proposed approach can also be combined with discrete-time optimal control strategies, for instance to achieve stabilization and tracking for robot systems on $\text{SE}(3)$. Possible future directions include extensions to multi-link robots and multi-agent  systems (e.g. on $(\text{SE}(3))^n$).

\newpage



\acks{The authors gratefully acknowledge support from NSF under grants CCF-2112665, DMS-1345013, DMS-1813635 and from AFOSR under grant FA9550-18-1-0288. }

\bibliography{bib/Bibliography.bib}

\begin{thebibliography}{53}
\providecommand{\natexlab}[1]{#1}
\providecommand{\url}[1]{\texttt{#1}}
\expandafter\ifx\csname urlstyle\endcsname\relax
  \providecommand{\doi}[1]{doi: #1}\else
  \providecommand{\doi}{doi: \begingroup \urlstyle{rm}\Url}\fi

\bibitem[Acosta et~al.(2014)Acosta, Sanchez, and Ollero]{acosta2014robust}
J.~A. Acosta, M.~I. Sanchez, and A.~Ollero.
\newblock {Robust control of underactuated aerial manipulators via IDA-PBC}.
\newblock In \emph{IEEE Conference on Decision and Control (CDC)}, 2014.

\bibitem[Amos et~al.(2018)Amos, Jimenez, Sacks, Boots, and
  Kolter]{amos2018differentiable}
B.~Amos, I.~Jimenez, J.~Sacks, B.~Boots, and J.~Z. Kolter.
\newblock {Differentiable MPC for End-to-end Planning and Control}.
\newblock In \emph{{Advances in Neural Information Processing Systems}}, 2018.

\bibitem[Bertalan et~al.(2019)Bertalan, Dietrich, Mezi{\'c}, and
  Kevrekidis]{Bertalan2019}
T.~Bertalan, F.~Dietrich, I.~Mezi{\'c}, and I.~G. Kevrekidis.
\newblock On learning {H}amiltonian systems from data.
\newblock \emph{Chaos: An Interdisciplinary Journal of Nonlinear Science},
  29\penalty0 (12):\penalty0 121107, 2019.
\newblock \doi{10.1063/1.5128231}.

\bibitem[Blanes and Casas(2017)]{Blanes2017}
S.~Blanes and F.~Casas.
\newblock \emph{A Concise Introduction to Geometric Numerical Integration}.
\newblock 2017.
\newblock ISBN 9781482263442.
\newblock \doi{10.1201/b21563}.

\bibitem[Borrelli et~al.(2017)Borrelli, Bemporad, and
  Morari]{borrelli_MPC_book}
F.~Borrelli, A.~Bemporad, and M.~Morari.
\newblock \emph{{Predictive control for linear and hybrid systems}}.
\newblock Cambridge University Press, 2017.

\bibitem[Burby et~al.(2020)Burby, Tang, and Maulik]{BurbyHenon}
J.~W. Burby, Q.~Tang, and R.~Maulik.
\newblock Fast neural {P}oincar{\'{e}} maps for toroidal magnetic fields.
\newblock \emph{Plasma Physics and Controlled Fusion}, 63\penalty0
  (2):\penalty0 024001, 2020.
\newblock \doi{10.1088/1361-6587/abcbaa}.

\bibitem[Chen et~al.(2021)Chen, Matsubara, and Yaguchi]{Chen2021neural}
Y.~Chen, T.~Matsubara, and T.~Yaguchi.
\newblock Neural symplectic form: learning {H}amiltonian equations on general
  coordinate systems.
\newblock In \emph{Advances in Neural Information Processing Systems}, 2021.

\bibitem[{Chen} et~al.(2020){Chen}, {Zhang}, {Arjovsky}, and
  {Bottou}]{Chen2020}
Z.~{Chen}, J.~{Zhang}, M.~{Arjovsky}, and L.~{Bottou}.
\newblock {Symplectic Recurrent Neural Networks}.
\newblock \emph{International Conference on Learning Representations}, 2020.

\bibitem[Cieza and Reger(2019)]{cieza2019ida}
O.~B. Cieza and J.~Reger.
\newblock {IDA-PBC for underactuated mechanical systems in implicit
  {P}ort-{H}amiltonian representation}.
\newblock In \emph{European Control Conference (ECC)}, 2019.

\bibitem[Cranmer et~al.(2020)Cranmer, Greydanus, Hoyer, Battaglia, Spergel, and
  Ho]{Cranmer2020}
M.~Cranmer, S.~Greydanus, S.~Hoyer, P.~W. Battaglia, D.~N. Spergel, and S.~Ho.
\newblock Lagrangian neural networks.
\newblock \emph{ICLR 2020 Workshop on Integration of Deep Neural Models and
  Differential Equations}, 2020.

\bibitem[Duong and Atanasov(2021)]{duong21hamiltonian}
T.~Duong and N.~Atanasov.
\newblock {Hamiltonian-based Neural ODE Networks on the SE(3) Manifold For
  Dynamics Learning and Control}.
\newblock In \emph{Proceedings of Robotics: Science and Systems}, 2021.
\newblock \doi{10.15607/RSS.2021.XVII.086}.

\bibitem[Duong and Atanasov(2022)]{duong2022adaptive}
T.~Duong and N.~Atanasov.
\newblock Adaptive control of {SE}(3) {H}amiltonian dynamics with learned
  disturbance features.
\newblock \emph{IEEE Control Systems Letters}, 2022.

\bibitem[Duruisseaux and Leok(2022)]{Duruisseaux2022Constrained}
V.~Duruisseaux and M.~Leok.
\newblock Accelerated optimization on {R}iemannian manifolds via discrete
  constrained variational integrators.
\newblock \emph{Journal of Nonlinear Science}, 32\penalty0 (42), 2022.

\bibitem[Duruisseaux and Leok(2023)]{Duruisseaux2022Lagrangian}
V.~Duruisseaux and M.~Leok.
\newblock Time-adaptive {L}agrangian variational integrators for accelerated
  optimization on manifolds.
\newblock \emph{Journal of Geometric Mechanics}, 15\penalty0 (1):\penalty0
  224--255, 2023.
\newblock ISSN 1941-4889.

\bibitem[Duruisseaux et~al.(2021)Duruisseaux, Schmitt, and
  Leok]{duruisseaux2020adaptive}
V.~Duruisseaux, J.~Schmitt, and M.~Leok.
\newblock Adaptive {H}amiltonian variational integrators and applications to
  symplectic accelerated optimization.
\newblock \emph{SIAM Journal on Scientific Computing}, 43\penalty0
  (4):\penalty0 A2949--A2980, 2021.

\bibitem[Duruisseaux et~al.(2023)Duruisseaux, Burby, and
  Tang]{DuruisseauxNPMap}
V.~Duruisseaux, J.~W. Burby, and Q.~Tang.
\newblock Approximation of nearly-periodic symplectic maps via
  structure-preserving neural networks.
\newblock \emph{Scientific Reports, Collection on ``Physics-informed Machine
  Learning and its real-world applications"}, 2023.
\newblock \doi{10.1038/s41598-023-34862-w}.

\bibitem[Gallier and Quaintance(2020)]{Gallier2020}
J.~Gallier and J.~Quaintance.
\newblock \emph{Differential Geometry and {L}ie Groups: A Computational
  Perspective}.
\newblock Geometry and Computing. Springer International Publishing, 2020.
\newblock ISBN 9783030460402.

\bibitem[Greydanus et~al.(2019)Greydanus, Dzamba, and Yosinski]{Greydanus2019}
S.~Greydanus, M.~Dzamba, and J.~Yosinski.
\newblock Hamiltonian neural networks.
\newblock In \emph{Advances in Neural Information Processing Systems},
  volume~32, 2019.

\bibitem[Gr{\"u}ne and Pannek(2017)]{NMPC_book}
L.~Gr{\"u}ne and J.~Pannek.
\newblock \emph{{Nonlinear model predictive control}}.
\newblock Springer, 2017.

\bibitem[Hairer et~al.(2006)Hairer, Lubich, and Wanner]{HaLuWa2006}
E.~Hairer, C.~Lubich, and G.~Wanner.
\newblock \emph{Geometric {N}umerical {I}ntegration}, volume~31 of
  \emph{Springer Series in Computational Mathematics}.
\newblock Springer-Verlag, 2006.

\bibitem[Havens and Chowdhary(2021)]{Havens2021}
A.~Havens and G.~Chowdhary.
\newblock Forced variational integrator networks for prediction and control of
  mechanical systems.
\newblock \emph{arXiv preprint arXiv:2106.02973}, 2021.

\bibitem[Holm et~al.(2009)Holm, Schmah, and Stoica]{Holm2009}
D.~Holm, T.~Schmah, and C.~Stoica.
\newblock \emph{Geometric Mechanics and Symmetry: From Finite to Infinite
  Dimensions}.
\newblock Oxford Texts in Applied and Engineering Mathematics. OUP Oxford,
  2009.
\newblock ISBN 9780199212910.

\bibitem[Jin et~al.(2020)Jin, Zhang, Zhu, Tang, and Karniadakis]{Jin2020}
P.~Jin, Z.~Zhang, A.~Zhu, Y.~Tang, and G.~E. Karniadakis.
\newblock {SympNets: Intrinsic structure-preserving symplectic networks for
  identifying Hamiltonian systems}.
\newblock \emph{Neural Networks}, 132\penalty0 (C), 12 2020.
\newblock \doi{10.1016/j.neunet.2020.08.017}.

\bibitem[Kingma and Ba(2014)]{ADAM}
D.~Kingma and J.~Ba.
\newblock Adam: A method for stochastic optimization.
\newblock In \emph{International Conference on Learning Representations}, 2014.

\bibitem[Lall and West(2006)]{LaWe2006}
S.~Lall and M.~West.
\newblock Discrete variational {H}amiltonian mechanics.
\newblock \emph{J. Phys. A}, 39\penalty0 (19):\penalty0 5509--5519, 2006.

\bibitem[Lee(2008)]{LeeThesis}
T.~Lee.
\newblock Computational geometric mechanics and control of rigid bodies.
\newblock \emph{Ph.D. dissertation, University of Michigan}, 2008.

\bibitem[Lee et~al.(2017)Lee, Leok, and McClamroch]{Lee2017}
T.~Lee, M.~Leok, and N.~H. McClamroch.
\newblock \emph{Global Formulations of {L}agrangian and {H}amiltonian Dynamics
  on Manifolds: A Geometric Approach to Modeling and Analysis}.
\newblock Interaction of Mechanics and Mathematics. Springer International
  Publishing, 2017.
\newblock ISBN 9783319569536.

\bibitem[Leimkuhler and Reich(2004)]{LeRe2005}
B.~Leimkuhler and S.~Reich.
\newblock \emph{Simulating {H}amiltonian Dynamics}, volume~14 of
  \emph{Cambridge Monographs on Applied and Computational Mathematics}.
\newblock Cambridge University Press, 2004.

\bibitem[Leok and Shingel(2012)]{LeSh2011}
M.~Leok and T.~Shingel.
\newblock Prolongation-collocation variational integrators.
\newblock \emph{IMA J. Numer. Anal.}, 32\penalty0 (3):\penalty0 1194--1216,
  2012.

\bibitem[Leok and Zhang(2011)]{LeZh2011}
M.~Leok and J.~Zhang.
\newblock Discrete {H}amiltonian variational integrators.
\newblock \emph{IMA Journal of Numerical Analysis}, 31\penalty0 (4):\penalty0
  1497--1532, 2011.

\bibitem[Lutter et~al.(2019{\natexlab{a}})Lutter, Listmann, and
  Peters]{lutter2019deepunderactuated}
M~Lutter, K~Listmann, and J~Peters.
\newblock {Deep Lagrangian Networks for end-to-end learning of energy-based
  control for under-actuated systems}.
\newblock In \emph{IEEE/RSJ International Conference on Intelligent Robots and
  Systems (IROS)}, 2019{\natexlab{a}}.

\bibitem[Lutter et~al.(2019{\natexlab{b}})Lutter, Ritter, and
  Peters]{Lutter2018}
M.~Lutter, C.~Ritter, and J.~Peters.
\newblock Deep {L}agrangian networks: Using physics as model prior for deep
  learning.
\newblock In \emph{International Conference on Learning Representations},
  2019{\natexlab{b}}.

\bibitem[Marco and M{\'e}hats(2021)]{Marco2021}
D.~Marco and F.~M{\'e}hats.
\newblock Symplectic learning for {H}amiltonian neural networks.
\newblock \emph{arXiv preprint arXiv:2106.11753}, 2021.

\bibitem[Marsden and Ratiu(1999)]{MaRa1999}
{J. E.} Marsden and {T. S.} Ratiu.
\newblock \emph{Introduction to mechanics and symmetry}, volume~17 of
  \emph{Texts in Applied Mathematics}.
\newblock Springer-Verlag, New York, second edition, 1999.

\bibitem[Marsden and West(2001)]{MaWe2001}
J.~E. Marsden and M.~West.
\newblock Discrete mechanics and variational integrators.
\newblock \emph{Acta Numer.}, 10:\penalty0 357--514, 2001.

\bibitem[Mathiesen et~al.(2022)Mathiesen, Yang, and Hu]{Mathiesen2022}
F.~B. Mathiesen, B.~Yang, and J.~Hu.
\newblock Hyperverlet: A symplectic hypersolver for {H}amiltonian systems.
\newblock \emph{Proceedings of the AAAI Conference on Artificial Intelligence},
  36\penalty0 (4):\penalty0 4575--4582, June 2022.
\newblock \doi{10.1609/aaai.v36i4.20381}.

\bibitem[Ober-Blöbaum et~al.(2011)Ober-Blöbaum, Junge, and Marsden]{Ober2011}
S.~Ober-Blöbaum, O.~Junge, and J.~E. Marsden.
\newblock Discrete mechanics and optimal control: An analysis.
\newblock \emph{ESAIM: Control, Optimisation and Calculus of Variations},
  17\penalty0 (2):\penalty0 322–352, 2011.
\newblock \doi{10.1051/cocv/2010012}.

\bibitem[Offen and Ober-Bl{\"o}baum(2022)]{Offen2022}
C.~Offen and S.~Ober-Bl{\"o}baum.
\newblock Symplectic integration of learned {H}amiltonian systems.
\newblock \emph{Chaos: An Interdisciplinary Journal of Nonlinear Science},
  32\penalty0 (1):\penalty0 013122, 2022.
\newblock \doi{10.1063/5.0065913}.

\bibitem[Ortega et~al.(2002)Ortega, Spong, G{\'o}mez-Estern, and
  Blankenstein]{ortega2002stabilization}
R.~Ortega, M.~W. Spong, F.~G{\'o}mez-Estern, and G.~Blankenstein.
\newblock Stabilization of a class of underactuated mechanical systems via
  interconnection and damping assignment.
\newblock \emph{IEEE Transactions on Automatic Control}, 47\penalty0 (8), 2002.

\bibitem[Panerati et~al.(2020)Panerati, Zheng, Zhou, Xu, Prorok, and
  Sch\"{o}llig]{gym-pybullet-drones2020}
J.~Panerati, H.~Zheng, S.~Zhou, J.~Xu, A.~Prorok, and A.~P. Sch\"{o}llig.
\newblock Learning to fly: a {PyBullet} gym environment to learn the control of
  multiple nano-quadcopters.
\newblock \url{https://github.com/utiasDSL/gym-pybullet-drones}, 2020.

\bibitem[Poincar\'e(1899)]{Poincare1899}
H.~Poincar\'e.
\newblock \emph{{Les m\'ethodes nouvelles de la m\'ecanique c\'eleste, Volume
  3}}.
\newblock Gauthier-Villars, Paris, 1899.

\bibitem[Rath et~al.(2021)Rath, Albert, Bischl, and von Toussaint]{Rath2021}
K.~Rath, C.~G. Albert, B.~Bischl, and U.~von Toussaint.
\newblock Symplectic {G}aussian process regression of maps in {H}amiltonian
  systems.
\newblock \emph{Chaos: An Interdisciplinary Journal of Nonlinear Science},
  31\penalty0 (5):\penalty0 053121, 2021.
\newblock \doi{10.1063/5.0048129}.

\bibitem[S{\ae}mundsson et~al.(2020)S{\ae}mundsson, Terenin, Hofmann, and
  Deisenroth]{Saemundsson2020}
S.~S{\ae}mundsson, A.~Terenin, K.~Hofmann, and M.~P. Deisenroth.
\newblock Variational integrator networks for physically structured embeddings.
\newblock In \emph{AISTATS}, 2020.

\bibitem[Santos et~al.(2022)Santos, Ekal, and Ventura]{Santos2022}
S.~Santos, M.~Ekal, and R.~Ventura.
\newblock Symplectic momentum neural networks - using discrete variational
  mechanics as a prior in deep learning.
\newblock In \emph{Learning for Dynamics and Control Conference}, pages
  584--595, 2022.

\bibitem[Schmitt and Leok(2017)]{ScLe2017}
J.~M. Schmitt and M.~Leok.
\newblock Properties of {H}amiltonian variational integrators.
\newblock \emph{IMA Journal of Numerical Analysis}, 38\penalty0 (1):\penalty0
  377--398, 03 2017.

\bibitem[Schmitt et~al.(2018)Schmitt, Shingel, and Leok]{ScShLe2017}
J.~M. Schmitt, T.~Shingel, and M.~Leok.
\newblock {L}agrangian and {H}amiltonian {T}aylor variational integrators.
\newblock \emph{{BIT} Numerical Mathematics}, 58:\penalty0 457--488, 2018.
\newblock \doi{10.1007/s10543-017-0690-9}.

\bibitem[Tassa et~al.(2014)Tassa, Mansard, and Todorov]{tassa2014control}
Y.~Tassa, N.~Mansard, and E.~Todorov.
\newblock Control-limited differential dynamic programming.
\newblock In \emph{IEEE International Conference on Robotics and Automation
  (ICRA)}, pages 1168--1175, 2014.

\bibitem[Valperga et~al.(2022)Valperga, Webster, Turaev, Klein, and
  Lamb]{Valperga2022}
R.~Valperga, K.~Webster, D.~Turaev, V.~Klein, and J.~Lamb.
\newblock Learning reversible symplectic dynamics.
\newblock In \emph{Learning for Dynamics and Control Conference}, volume 168,
  pages 906--916. PMLR, 2022.

\bibitem[Van Der~Schaft and Jeltsema(2014)]{van2014port}
A.~Van Der~Schaft and D.~Jeltsema.
\newblock Port-{H}amiltonian systems theory: An introductory overview.
\newblock \emph{Foundations and Trends in Systems and Control}, 1\penalty0
  (2-3), 2014.

\bibitem[Willard et~al.(2020)Willard, Jia, Xu, Steinbach, and
  Kumar]{Willard2020}
J.~D. Willard, X.~Jia, S.~Xu, M.~S. Steinbach, and V.~Kumar.
\newblock Integrating physics-based modeling with machine learning: A survey.
\newblock \emph{arXiv preprint arXiv:2003.04919}, 2020.

\bibitem[Zhong et~al.(2020{\natexlab{a}})Zhong, Dey, and
  Chakraborty]{Zhong2020}
Y.~D. Zhong, B.~Dey, and A.~Chakraborty.
\newblock {Symplectic ODE-Net: Learning Hamiltonian} dynamics with control.
\newblock In \emph{International Conference on Learning Representations},
  2020{\natexlab{a}}.

\bibitem[Zhong et~al.(2020{\natexlab{b}})Zhong, Dey, and
  Chakraborty]{zhong2020dissipative}
Y.~D. Zhong, B.~Dey, and A.~Chakraborty.
\newblock Dissipative {S}ym{ODEN}: Encoding {H}amiltonian dynamics with
  dissipation and control into deep learning.
\newblock In \emph{ICLR 2020 Workshop on Integration of Deep Neural Models and
  Differential Equations}, 2020{\natexlab{b}}.

\bibitem[Zhong et~al.(2021)Zhong, Dey, and Chakraborty]{Zhong2021}
Y.~D. Zhong, B.~Dey, and A.~Chakraborty.
\newblock Benchmarking energy-conserving neural networks for learning dynamics
  from data.
\newblock In \emph{Learning for Dynamics and Control}, volume 144, pages
  1218--1229. PMLR, 2021.

\end{thebibliography}

\appendix

\newpage 

\section{Rigid-body kinematics on $\text{SE}(3)$} 
\label{appendix: Rigid-Body Kinematics}


We present here a brief introduction to rigid-body kinematics on $\text{SE}(3)$, mostly extracted from Chapters 2, 6, 7 of \citet{Lee2017}. \\
 
A rigid body is an idealization of a real mechanical system, defined as a collection of material particles such that the relative distance between any two particles in the body does not change (i.e, the body does not deform). The configuration of a rigid body is a representation of its position and attitude in 3-dimensional space. The kinematics of a rigid body describe how its configuration changes under the influence of linear velocity and angular velocity. Defining the configuration of the rigid body is of the utmost importance for rigid-body kinematics, and depends on the constraints imposed on the rigid-body motion.

\subsection{Rotational Rigid-Body Motion}


If a rigid body has fixed position but can rotate arbitrarily in $\mathbb{R}^3$, then its configuration can be represented by a rotation matrix. Hence, the manifold of rotation matrices, $\text{SO}(3)$, is the configuration manifold for rigid-body rotational motion. Since the dimension of $\text{SO}(3)$ is three, rigid-body rotations have three degrees of freedom. \\

We use two Euclidean frames: an arbitrary reference frame and another frame fixed to the rigid body which rotates with it (with origin selected at the center of mass of the rigid body). A rotation matrix $R \in  \text{SO}(3)$ is a linear transformation on $\mathbb{R}^3$ between the body-fixed and reference frames:
\begin{itemize}
     \item if $v \in \mathbb{R}^3$ represents a vector in the body frame, then $Rv \in \mathbb{R}^3$ represents the same vector in the reference frame,
     \item if $v \in \mathbb{R}^3$ represents a vector in the reference frame, then $R^\top v \in \mathbb{R}^3$ represents the same vector in the body frame.
\end{itemize}
We can describe the rotation of the rigid body through the rotation of the body-fixed frame: the configuration of a rotating rigid body is the linear transformation that relates the representation of a vector in the body-fixed frame to its representation in the reference frame. 

Suppose that $R(t) \in \text{SO}(3)$ represents the rotational motion of a rigid body. Differentiating the orthogonality condition $R^\top R = \mathbb{I}_3$, we get $\dot{R}^\top R = - R^\top \dot{R}$ which implies that $R^\top \dot{R}$ remains skew-symmetric at all time. Thus, there exists a skew-symmetric matrix $\xi(t) \in \mathfrak{so}(3)$ such that $R^\top \dot{R} = \xi$, from which we can obtain the rotational kinematics:
\begin{equation}
    \dot{R} = R \xi.
\end{equation}
Using the isomorphism between the Lie algebra $\mathfrak{so}(3)$ and $\mathbb{R}^3$ given by $\xi = S(\omega)$ for $\omega\in \mathbb{R}^3$ and $\xi \in \mathfrak{so}(3)$, we can rewrite the rotational kinematics as 
\begin{equation}
    \dot{R} = R S(\omega),
\end{equation}
where $\omega \in \mathbb{R}^3$ is referred to as the angular velocity vector of the rigid body expressed in the body frame. Thus, the rotational kinematics describe the rate of change $\dot{R}$ of the configuration in terms of the angular velocity $\omega \in \mathbb{R}^3$ represented in the body frame.

\subsection{General Rigid-Body Motion}

\hfill 
\vspace{-2mm}

General rigid-body motion can be described by a combination of rotations and translations. \\

As before, we use two inertial frames: a first arbitrary reference frame and another frame fixed to the rigid body which translates and rotates with the rigid body (with origin usually selected at the center of mass of the rigid body). The translational configuration of the rigid body characterizes the motion of the body-fixed frame origin and can be selected to lie in the configuration manifold $\mathbb{R}^3$. \\

The configuration manifold for a rigid-body that is simultaneously translating and rotating can be selected as the semidirect product of $\mathbb{R}^3$ and $\text{SO}(3)$. Therefore, we can represent the configuration via $(R, x) \in \text{SE}(3)$ in the sense that $R \in \text{SO}(3)$ is the orientation and $x \in \mathbb{R}^3$ is the position of the body-fixed frame in the reference frame. Consequently, the Lie group $\text{SE}(3)$ can be viewed as the configuration manifold for general rigid-body motion. \\

As before, the rotational kinematics describe the rate of change $\dot{R}$ of the configuration in terms of the angular velocity vector $\omega \in \mathbb{R}^3$ of the rigid body represented in the body-fixed frame: 
\begin{equation}
    \dot{R} = R S(\omega).
\end{equation}
Now, the translational velocity vector $v \in \mathbb{R}^3$ of the rigid body (i.e., of the origin of the body-fixed frame) is the time derivative of the position vector from the origin of the reference frame to the origin of the body-fixed frame. In the reference frame, the
translational velocity vector $\dot{x} \in \mathbb{R}^3$ of the rigid body is
\begin{equation} \dot{x} = Rv. \end{equation}
These are referred to as the translational kinematics of the rigid body. Altogether, we obtain the kinematics for general rigid-body motion:
\begin{equation}
    \dot{R} = R S(\omega), \qquad  \quad  \dot{x} = Rv.
\end{equation}

\hfill \\

\section{Derivation of the forced variational integrator on $\text{SE}(3)$} \label{appendix: Derivation of integrator}

\hfill  
\vspace{-2mm}

In this appendix, we will derive the forced discrete Euler--Lagrange equations in Lagrangian form (equations~\eqref{eq: discrete EL 1}-\eqref{eq: discrete EL 3}) and in Hamiltonian form (equations~\eqref{eq: SE3 discrete H equation 2}-\eqref{eq: SE3 discrete H equation 4}) associated to the discrete Lagrangian $L_d$ and discrete control forces $\mathcal{f}_d^{\pm}$ on $\text{SE}(3)$ presented in Section~\ref{sec: Variational Integrator}. \\

Consider a Lie group $G$  with associated Lie algebra $\mathfrak{g} = T_eG$. In what follows, $\text{L} : G\times G \rightarrow G$ denotes the left action on $G$, defined by $\text{L}_q h = qh$ for all $q,h\in G$. The adjoint operator is denoted by $\text{Ad}_q : \mathfrak{g} \rightarrow \mathfrak{g},$ and $\text{Ad}^*_q : \mathfrak{g}^* \rightarrow \mathfrak{g}^*$ denotes the corresponding coadjoint. We refer the reader to~\citep{MaRa1999,Lee2017,Gallier2020} for a more detailed description of Lie group theory and mechanics on Lie groups.  \\

Given a discrete Lagrangian $L_d(g_k,z_k)$ on the Lie group $G$, the forced discrete Euler--Lagrange equations are given by
	\begin{equation}
	   g_{k+1} = g_k \star z_k,
	\end{equation}
	\begin{equation}  \text{T}_e^* \text{L}_{z_{k-1}}   D_2 L_{d_{k-1}}   - \text{Ad}^*_{z_k^{-1}} \left( \text{T}_e^* \text{L}_{z_k}   D_2 L_{d_k}   \right)  +  \text{T}_e^* \text{L}_{g_k}  D_1  L_{d_k}  +  \mathcal{f}_{d_k}^- + \mathcal{f}_{d_{k-1}}^+ = 0    ,    
	\end{equation}  
	where $L_{d_k} = L_d(g_k,z_k)$ and $\mathcal{f}_{d_k}^{\pm} = \mathcal{f}_d^{\pm}(g_k, g_{k+1}, u_k)$. \\
	
\noindent 	Using the discrete Legendre transform
\begin{align} \mu_{k} & =  \text{Ad}^*_{z_k^{-1}} \left( \text{T}_e^* \text{L}_{z_k}   D_2 L_{d_k}    \right) -  \text{T}_e^* \text{L}_{g_k}  D_1  L_{d_k} - \mathcal{f}_{d_k}^- ,  \end{align} 
we can rewrite the equations of motion in Hamiltonian form as
	\begin{equation}\mu_{k}  =  \text{Ad}^*_{z_k^{-1}} \left( \text{T}_e^* \text{L}_{z_k}   D_2 L_{d_k}    \right) -  \text{T}_e^* \text{L}_{g_k}  D_1  L_{d_k}  - \mathcal{f}_{d_k}^- , \end{equation}  \begin{equation}
    \mu_{k+1}  =  \text{T}_e^* \text{L}_{z_k}   D_2 L_{d_k} +  \mathcal{f}_{d_k}^+ =\text{Ad}^*_{z_k} (\mu_k + \text{T}_e^* \text{L}_{g_k}  D_1  L_{d_k}  +  \mathcal{f}_{d_k}^-  ) +  \mathcal{f}_{d_k}^+  ,
\end{equation}
	\begin{equation}
	   g_{k+1} = g_k \star z_k  .
	\end{equation}

\hfill 

On $\text{SE}(3)$, with $g_k = (x_k,R_k) \in \text{SE}(3)$ and $z_k = (y_k, Z_k)\in \text{SE}(3)$, the discrete kinematics equations $g_{k+1} = g_k \star z_k$ are given by 
\begin{equation}
    R_{k+1} = R_k Z_k  \qquad \text{ and } \qquad x_{k+1} = x_k +R_k y_k,
\end{equation}
so that $\{(x_k, R_k) \}$ remains on $\text{SE}(3)$. Using the kinematics equation $\dot{R} = RS(\omega)$, the matrix $S(\omega_k)$ can be approximated via
\begin{equation}
    S(\omega_k) = R_k^\top \dot{R}_k \approx R_k^\top \frac{R_{k+1} - R_k}{h}  = \frac{1}{h} ( Z_k - \mathbb{I}_3).
\end{equation}

\hfill 

\noindent With the discrete Lagrangian
\begin{equation}  
\begin{aligned}
L_d(x_k, R_k, y_k, Z_k  )  &  =  \frac{m}{2h} y_k^\top y_k + \frac{1}{h} \text{tr}\left( [\mathbb{I}_3  - Z_k] J_d \right)  \\ &  \qquad \qquad   - (1-\alpha ) h U(x_k, R_k) - \alpha  h  U(x_k + R_ky_k, R_k Z_k) ,
\end{aligned}
\end{equation}

\noindent it can be shown by proceeding as in~\citep{LeeThesis} that the forced discrete Euler--Lagrange equations are given by
\begin{equation}
    \frac{1}{h}( J_d Z_{k-1}- Z_{k-1}^\top J_d ) - \frac{1}{h}(Z_k J_d - J_d Z_k^\top ) +  h S(\xi_k) + S( \mathcal{f}_{d_k}^{R-} ) + S( \mathcal{f}_{d_{k-1}}^{R+}  )  = 0 ,
\end{equation}
\begin{equation}
    \frac{m}{h} R_{k}^\top  (x_k - x_{k-1}) - \frac{m}{h} R_k^\top (x_{k+1} - x_k)-  hR_k^\top \frac{\partial U_k}{\partial x_k} +   \mathcal{f}_{d_k}^{x-} + \mathcal{f}_{d_{k-1}}^{x+}   = 0,
\end{equation}
\begin{equation}
    R_{k+1} = R_k Z_k,
\end{equation}
where $\mathcal{f}_{d_k}^{x \pm}$ and $\mathcal{f}_{d_k}^{R \pm}$ denote the $x$ and $R$ components of the discrete forces $\mathcal{f}_{d_k}^{\pm}$. \\

\noindent This can be simplified into the forced discrete Euler--Lagrange equations
\begin{equation} 
       h^2 S(\xi_k) + h S( \mathcal{f}_{d_k}^{R-} ) + h S( \mathcal{f}_{d_{k-1}}^{R+}  )  + ( J_d  Z_{k-1} - Z_{k-1}^\top  J_d )   =  Z_k J_d  - J_d Z_k^\top  ,
\end{equation}
\begin{equation} 
 x_{k+1} = 2x_k - x_{k-1} - \frac{h^2}{m} \frac{\partial U_k}{\partial x_k} + \frac{h}{m} R_{k} ( \mathcal{f}_{d_k}^{x-} - \mathcal{f}_{d_{k-1}}^{x+} ) ,  
\end{equation}
\begin{equation}
 R_{k+1} = R_k Z_k.
\end{equation}

\hfill 

\noindent Using the discrete Legendre transforms
\begin{equation}
    S(\pi_k) = \frac{1}{h}( Z_k J_d - J_d Z_k^\top  ) - (1-\alpha ) h S(\xi_k)  -  S( \mathcal{f}_{d_k}^{R-} ),
\end{equation}
\begin{equation}
    \nu_k = \frac{m}{h}R_k^\top (x_{k+1}- x_{k})  + (1-\alpha ) h R_k^\top \frac{\partial U_k}{\partial x_k} -  \mathcal{f}_{d_k}^{x-},
\end{equation}
we get
\begin{equation}
    S(\pi_{k+1}) = \frac{1}{h}(  J_d Z_k - Z_k^\top J_d  ) + \alpha h S(\xi_{k+1}) +  S( \mathcal{f}_{d_k}^{R+}),
\end{equation}
\begin{equation} \label{eq: appendix gamma k+1}
    \nu_{k+1} = \frac{m}{h} R_{k+1}^\top (x_{k+1}- x_{k})  - \alpha h R_{k+1}^\top \frac{\partial U_{k+1}}{\partial x_{k+1}} +  \mathcal{f}_{d_k}^{x+}.
\end{equation}

\hfill  \\

\noindent With $\gamma = R\nu$, equation~\eqref{eq: appendix gamma k+1} can be rewritten as
\begin{equation}
    \gamma_{k+1} = \frac{m}{h}  (x_{k+1}- x_{k})  - \alpha h \frac{\partial U_{k+1}}{\partial x_{k+1}} + R_{k+1} \mathcal{f}_{d_k}^{x+} .
\end{equation}

\hfill

\noindent Overall, we obtain the following implicit discrete equations of motion in Hamiltonian form: 
\begin{align}
    &   S(\pi_k) = \frac{1}{h}( Z_k J_d - J_d Z_k^\top  ) - (1-\alpha ) h S(\xi_k)  -  S(\mathcal{f}_{d_k}^{R-}) \label{eq: Appendix H1} , \\ 
    &  \gamma_k = \frac{m}{h} (x_{k+1}- x_{k})  + (1-\alpha ) h  \frac{\partial U_k}{\partial x_k} -  R_k \mathcal{f}_{d_k}^{x-} \label{eq: Appendix H2} , \\ 
    &   R_{k+1} = R_k Z_k , \\
    &  S(\pi_{k+1}) = \frac{1}{h}(  J_d Z_k - Z_k^\top J_d  ) + \alpha h S(\xi_{k+1}) +  S( \mathcal{f}_{d_k}^{R+}) \label{eq: Appendix H3} , \\
     & \gamma_{k+1} = \frac{m}{h}  (x_{k+1}- x_{k})  - \alpha  h
     \frac{\partial U_{k+1}}{\partial x_{k+1}} + R_{k+1} \mathcal{f}_{d_k}^{x+}  \label{eq: Appendix H4} . 
\end{align}

\hfill 

\noindent Equations~\eqref{eq: Appendix H1} and~\eqref{eq: Appendix H2} give
\begin{equation}
    hS(\pi_k) + (1-\alpha ) h^2  S(\xi_k) =  Z_k J_d - J_d Z_k^\top   -    h S( \mathcal{f}_{d_k}^{R-}),
\end{equation}
\begin{equation}
    x_{k+1} =  x_{k} + \frac{h}{m} \gamma_k - (1-\alpha ) \frac{ h^2}{m} \frac{\partial U_k}{\partial x_k} - \frac{h}{m} R_k \mathcal{f}_{d_k}^{x-}.
\end{equation}

\hfill 

\noindent Equation~\eqref{eq: Appendix H3} can be rewritten using equation~\eqref{eq: Appendix H1} as
\begin{equation}
     S(\pi_{k+1}) = Z_k^\top S(\pi_k) Z_k + (1-\alpha ) h  Z_k^\top S(\xi_k) Z_k + \alpha h S( \xi_{k+1})   + Z_k^\top S(\mathcal{f}_{d_k}^{R-}) Z_k + S(\mathcal{f}_{d_k}^{R+}).
\end{equation}

\hfill 

\noindent Since $Z^\top S(\eta) Z = S(Z^\top \eta)$ for any $Z \in \text{SO}(3)$ and $\eta \in \mathfrak{so}(3)$, we get
\begin{equation}
     \pi_{k+1}  = Z_k^\top \pi_k + (1-\alpha ) h Z_k^\top \xi_k+ \alpha h \xi_{k+1}   + Z_k^\top \mathcal{f}_{d_k}^{R-}+ \mathcal{f}_{d_k}^{R+}  .
\end{equation}

\hfill 

\noindent Finally, equation~\eqref{eq: Appendix H4} can be rewritten using equation~\eqref{eq: Appendix H2} as
\begin{equation}
     \gamma_{k+1} = \gamma_k - (1-\alpha ) h    \frac{\partial U_k}{\partial x_k} - \alpha h \frac{\partial U_{k+1}}{\partial x_{k+1}} +  R_k \mathcal{f}_{d_k}^{x-} + R_{k+1} \mathcal{f}_{d_k}^{x+} .
\end{equation}
   
\hfill \\

\noindent Overall, this gives the forced variational integrator~\eqref{eq: SE3 discrete H equation 2}-\eqref{eq: SE3 discrete H equation 4}:
\begin{align}
    &   hS(\pi_k) + (1-\alpha ) h^2 S(\xi_k) =  Z_k J_d - J_d Z_k^\top   -    h S( \mathcal{f}_{d_k}^{R-}) ,\\
    &   R_{k+1} = R_k Z_k , \\ 
    &  \pi_{k+1}  = Z_k^\top \pi_k + (1-\alpha ) h Z_k^\top \xi_k+ \alpha h \xi_{k+1}   + Z_k^\top \mathcal{f}_{d_k}^{R-}+ \mathcal{f}_{d_k}^{R+} , \\ 
    &   x_{k+1} =  x_{k} + \frac{h}{m} \gamma_k - (1-\alpha ) \frac{ h^2}{m} \frac{\partial U_k}{\partial x_k} -  \frac{h}{m} R_k \mathcal{f}_{d_k}^{x-} , \\
    &   \gamma_{k+1} = \gamma_k - (1-\alpha ) h    \frac{\partial U_k}{\partial x_k} - \alpha  h \frac{\partial U_{k+1}}{\partial x_{k+1}} +  R_k \mathcal{f}_{d_k}^{x-} +R_{k+1} \mathcal{f}_{d_k}^{x+} .
\end{align}

\hfill \\

\section{Transforming the equation $S(a) = Z J_d  - J_d Z^\top$} \label{appendix: Equation Transformation}

\hfill 
\vspace{-1.5mm}

\noindent Plugging the Cayley transform 
\begin{equation}  Z = \text{Cay}(\mathcal{z}) \equiv  (\mathbb{I}_3+S(\mathcal{z}))(\mathbb{I}_3-S(\mathcal{z}))^{-1} , \end{equation}
into the equation
\begin{equation}  S(a) = Z J_d  - J_d Z^\top, \end{equation} and using the fact that $
    (\mathbb{I}_3 \pm S(\mathcal{z}))^\top = (\mathbb{I}_3 \mp S(\mathcal{z}))$ gives
\begin{equation}
    S(a) = (\mathbb{I}_3+S(\mathcal{z}))(\mathbb{I}_3-S(\mathcal{z}))^{-1}  J_d  - J_d (\mathbb{I}_3+S(\mathcal{z}))^{-1} (\mathbb{I}_3-S(\mathcal{z})) .
\end{equation}

\vspace{3mm}

\noindent Now, $(\mathbb{I}_3 \pm S(\mathcal{z}))$ and $(\mathbb{I}_3 \mp S(\mathcal{z}))^{-1}$ commute, so we can rewrite the previous equation as
\begin{equation} \label{eq: temp equation 1 in Cayley section}
    S(a) = (\mathbb{I}_3-S(\mathcal{z}))^{-1}(\mathbb{I}_3+S(\mathcal{z}))  J_d  - J_d (\mathbb{I}_3-S(\mathcal{z}))(\mathbb{I}_3+S(\mathcal{z}))^{-1}  .
\end{equation}

\noindent Multiplying both sides of equation~\eqref{eq: temp equation 1 in Cayley section} on the left by $(\mathbb{I}_3-S(\mathcal{z}))$ and on the right by $(\mathbb{I}_3+S(\mathcal{z}))$ gives
\begin{equation}
    (\mathbb{I}_3-S(\mathcal{z})) S(a) (\mathbb{I}_3+S(\mathcal{z}))= (\mathbb{I}_3+S(\mathcal{z}))  J_d  (\mathbb{I}_3+S(\mathcal{z}))  - (\mathbb{I}_3-S(\mathcal{z})) J_d (\mathbb{I}_3-S(\mathcal{z})),
\end{equation}
which can be simplified into
\begin{equation} \label{eq: temp equation 2 in Cayley section}
 S(a) - S(\mathcal{z})S(a) + S(a)S(\mathcal{z})  - S(\mathcal{z}) S(a) S(\mathcal{z})   =   2 S(\mathcal{z})J_d + 2 J_d S(\mathcal{z}).
\end{equation}

\vspace{2.5mm}

\noindent Using
$   S(\mathcal{z})J_d + J_dS(\mathcal{z}) = S(J\mathcal{z})$ and the general formulas
\begin{equation}
    -S(y)S(x) + S(x)S(y) = S(S(x)y), \qquad 
 S(x)S(y)S(x) = - (y^\top x)S(x),
\end{equation}
we can simplify equation~\eqref{eq: temp equation 2 in Cayley section} into
\begin{equation}
 S(a) + S(S(a)\mathcal{z}) + (a^\top \mathcal{z}) S(\mathcal{z})    =   2 S(J\mathcal{z}).
\end{equation} 

\vspace{1.5mm}

\noindent This can be rewritten in the desired vector form
\begin{equation}
    a + a \times \mathcal{z} + (a^\top \mathcal{z})\mathcal{z} - 2J\mathcal{z} = 0.
\end{equation}

\hfill

\section{Implementation details}
\label{appendix: Implementation}

\hfill 
\vspace{-2.5mm}

\noindent In this appendix, we provide additional details concerning the implementation of the LieFVINs for the planar pendulum on $\text{SO}(3)$ and for the crazyflie quadrotor on $\text{SE}(3)$. In particular, we detail the structure of the neural networks, the data generation process, and the training process. \\

To train the dynamics model with Algorithm Ia, we minimize the loss function
\begin{equation}
        \mathcal{L}_{\text{Ia}}(\theta) = \sum_{i=1}^N \Vert x_1 - \tilde{x}_1\Vert^2 + \left\Vert \log\left(\tilde{R}_{1} R_1^\top\right)^\vee \right\Vert ^2 + \Vert v_1 - \tilde{v}_1\Vert^2 + \Vert \omega_1 - \tilde{\omega}_1\Vert^2, \end{equation}
   while we use the following loss function for Algorithm Ib
        \begin{equation}
        \begin{aligned}
        \mathcal{L}_{\text{Ib}}(\theta) &= \sum_{i=1}^N \Vert x_1 - \tilde{x}_1\Vert^2 + \Vert v_1 - \tilde{v}_1\Vert^2 + \Vert \omega_1 - \tilde{\omega}_1\Vert^2  \\
        & \qquad \qquad \quad  +   \left\Vert h S(J\omega_0) + h S( \mathcal{f}_{d_0}^{R-}) + (1-\alpha)h^2 S(\xi_0) - J_d Z_0 + Z_0^\top J_d \right\Vert^2.
        \end{aligned}
\end{equation}
 The network parameters $\theta$ are updated using Adam \citep{ADAM}, where the gradients $\partial\mathcal{L}/\partial\theta$ are calculated by back-propagation.  \\
 
 In the descriptions of the network architectures below, the first number is the input dimension while the last number is the output dimension. The hidden layers are listed in-between with their dimensions and activation functions.

\subsection{Pendulum}
\label{appendix: Implementation Pendulum}

\hfill 
\vspace{-2.5mm}

\noindent We use neural networks to represent the inertial matrix $J(q) = L(q)L(q)^\top + \epsilon$, the potential energy $U(q)$ and the input gains $g(q)$ as follows:
\begin{itemize}
    \item $L(q)$: $\ $  9 - 10 Tanh - 10 Tanh - 10 Linear - 6
    \item $U(q)$: $\ $  9 - 10 Tanh - 10 Tanh - 10 Linear - 1
    \item $g(q)$: $\ $  9 - 10 Tanh - 10 Tanh - 10 Linear - 3
\end{itemize}

\noindent The training data of the form $\{(\cos{\varphi}, \sin{\varphi}, \dot{\varphi})\}$ was collected from an OpenAI Gym environment, provided by \citep{Zhong2020}. The control inputs are sampled in $[-3, 3]$ and applied to the planar pendulum for $10$ time intervals of $0.02$s to generate $512$ state-control trajectories. The $\text{SO}(3)$ LieFVIN, as described in Algorithm Ia with $\alpha = 0.5$, was trained with a fixed learning rate of $10^{-3}$ for $10000$ iterations. \\

For comparison, we also learned the dynamics using a black-box model which is a multilayer perceptron $\text{MLP}(q, \dot{q}, u)$ with architecture [22 - 1000 Tanh - 1000 Tanh - 1000 Linear - 18]. \\

To drive the pendulum from downward position $\varphi = 0$ to a stabilized upright position $\varphi^* = \pi$, $\dot{\varphi}^* = 0$, with input constraint $|u| \leq 20$, the running cost $\mathcal{C}_d$ and terminal cost $ \Phi_d$ in the MPC problem are chosen to be \begin{equation}  C_d(R_{\ell +k},\omega_{\ell+k}, u_{\ell+k} )  \  = \ \text{tr}(\mathbb{I}_3 - R^{*\top}R_{\ell +k}) + 0.1\Vert \omega_{\ell+k}\Vert^2 + 10^{-4}\Vert u_{\ell+k} \Vert^2, \end{equation}
\begin{equation}  \Phi_d(R_{\ell +k},\omega_{\ell+k}, u_{\ell+k} )  \  = \  \text{tr}(\mathbb{I}_3 - R^{*\top}R_{\ell +k}) + 0.1\Vert \omega_{\ell+k}\Vert^2 + 10^{-4}\Vert u_{\ell+k} \Vert^2. \end{equation} 

\hfill 

\vspace{6mm}

\subsection{Crazyflie Quadrotor}
\label{appendix: Implementation Quadrotor}

\hfill 
\vspace{-2.5mm}

\noindent We use neural networks to represent the mass $m = r^2$, inertial matrix $J(q) = LL^\top + \epsilon$, the potential energy $U(q)$ and the input gains $g(q) = \begin{bmatrix} g_x(q) & g_R(q) \end{bmatrix}$ as follows:
\begin{itemize}
    \item $r$: $\ $ 1D pytorch parameter
    \item $L$:  $\ $ $3\times 3$ upper-triangular parameter matrix
    \item $U(q)$: $\ $ 9 - 10 Tanh - 10 Tanh- 10 Tanh - 10 Linear - 1
    \item $g(q)$: $\ $ 9 - 10 Tanh - 10 Tanh- 10 Tanh - 10 Linear - 24
\end{itemize}

To obtain the training data, the quadrotor was controlled from a random starting point to $36$ different desired poses using a PID controller, yielding $36$ $4$-second trajectories. The trajectories were used to generate a dataset of $N=2700$ position-velocity updates $\left\{ \left(q_0, \dot{q}_0, u_0\right) \mapsto \left(q_1, \dot{q}_1\right)  \right\}$ with time step $0.02$s. The $\text{SE}(3)$ LieFVIN,  as described in Algorithm Ib with $\alpha = 0.5$, was trained with a decaying learning rate initialized at $5\times 10^{-3}$ for $20000$ iterations. \\

To track a diamond-shaped trajectory using the model learnt by the LieFVIN, with control input constraints $0\leq f \leq 0.595, \  \vert \tau \vert \leq 10^{-3}[5.9 \ \ 5.9 \ \ 7.4 ]^\top$, the running cost $\mathcal{C}_d$ and terminal cost $ \Phi_d$ in the MPC problem are chosen to be \begin{equation} \begin{aligned}  C_d(x_{\ell+k}, R_{\ell +k}, v_{\ell+k}, \omega_{\ell+k}, u_{\ell+k}) \ & = \ 1.2\Vert x_{\ell+k}\Vert^2 + 10^{-5} \ \text{tr}(\mathbb{I}_3 - R_{\ell +k})  + 1.2\Vert v_{\ell+k}\Vert^2 \\ & \qquad \quad  + 10^{-4}\Vert \omega_{\ell+k}\Vert^2 + 10^{-6}\Vert u_{\ell+k} \Vert^2 , \end{aligned}  \end{equation}
\begin{equation} \begin{aligned}  \Phi_d(x_{\ell+k}, R_{\ell +k}, v_{\ell+k}, \omega_{\ell+k}, u_{\ell+k}) \ & = \ 1.2\Vert x_{\ell+k}\Vert^2 + 10^{-5} \ \text{tr}(\mathbb{I}_3 - R_{\ell +k})  + 1.2\Vert v_{\ell+k}\Vert^2 \\ &  \qquad \quad + 10^{-4}\Vert \omega_{\ell+k}\Vert^2 + 10^{-6}\Vert u_{\ell+k} \Vert^2 . \end{aligned}  \end{equation}

\hfill  \\

\vspace{3mm}

\section{Learning and controlling Lagrangian systems from position data} \label{appendix: position only}

\vspace{3.5mm}

\subsection{Problem Statement}

\hfill 
\vspace{-3mm}

\noindent We now consider the problem of learning controlled Lagrangian dynamics only from  position data: given a position-only dataset of trajectories for a Lagrangian system, we wish to infer the update map that generates these trajectories, while preserving the symplectic structure underlying the dynamical system and constraining the updates to the Lie group on which it evolves. More precisely, we wish to solve the following problem:
\begin{problem}
 Given a dataset of position-only updates $\left\{ \left(q_0^{(i)}, q_1^{(i)}, u_0^{(i)}, u_1^{(i)}\right) \mapsto q_2^{(i)} \right\}_{i=1}^{N}$ for a controlled Lagrangian system evolving on the Lie group $\mathcal{Q}$, we wish to find a symplectic mapping $\Psi : \mathcal{Q} \times \mathcal{Q} \times \mathcal{U} \times \mathcal{U} \rightarrow \mathcal{Q} $ which minimizes \begin{equation}
     \sum_{i=1}^{N}{\mathcal{D}_{ \mathcal{Q}}\left(q_2^{(i)}, \Psi\left(q_0^{(i)}, q_1^{(i)}, u_0^{(i)}, u_1^{(i)}\right)  \right)} ,
 \end{equation}
 where $\mathcal{D}_{\mathcal{Q}}$ is a distance metric on $\mathcal{Q}$. \\
\end{problem}

\vspace{2mm} 

\subsection{Forced Variational Integrator in Lagrangian Form} \label{appendix: forced VI Lagrangian}

\hfill 
\vspace{-3mm}

\noindent As before, we choose the discrete Lagrangian
\begin{equation}  \label{eq: discrete Lagrangian appendix}
\begin{aligned}
L_d(x_k, R_k, y_k, Z_k  )  &  =  \frac{m}{2h} y_k^\top y_k + \frac{1}{h} \text{tr}\left( [\mathbb{I}_3  - Z_k] J_d \right)  \\ &  \qquad \qquad - (1-\alpha ) h U(x_k, R_k) - \alpha  h  U(x_k + R_ky_k, R_k Z_k),
\end{aligned}
\end{equation} 
where $\alpha \in [0,1]$ and $J_d = \frac{1}{2} \text{tr}(J) \mathbb{I}_3 - J$. We also define $U_k$ and $\xi_k$  via 
\begin{equation}
 U_k = U(x_k,R_k)   \quad \text{ and } \quad  S(\xi_k) = \frac{\partial  U_{k}}{\partial R_{k}}^\top R_{k} - R_{k}^\top \frac{\partial U_{k}}{\partial R_{k}} .
\end{equation}

\newpage 

It is shown in Appendix~\ref{appendix: Derivation of integrator} that the forced discrete Euler--Lagrange equations associated to the discrete Lagrangian~\eqref{eq: discrete Lagrangian appendix} and the discrete control forces $
    \mathcal{f}_{d_k}^{\pm} \equiv \mathcal{f}_{d}^{\pm}(x_k, R_k, u_k)$ are given by
    \begin{equation} \label{eq: discrete EL 1}
       h^2 S(\xi_k) + h S( \mathcal{f}_{d_k}^{R-} ) + h S( \mathcal{f}_{d_{k-1}}^{R+}  ) + ( J_d  Z_{k-1} - Z_{k-1}^\top  J_d )   =  Z_k J_d  - J_d Z_k^\top   ,
\end{equation}
\begin{equation} \label{eq: discrete EL 2}
 x_{k+1} = 2x_k - x_{k-1} - \frac{h^2}{m} \frac{\partial U_k}{\partial x_k} + \frac{h}{m} R_{k} ( \mathcal{f}_{d_k}^{x-} - \mathcal{f}_{d_{k-1}}^{x+} ) ,  
\end{equation}
\begin{equation}
    \label{eq: discrete EL 3} R_{k+1} = R_k Z_k.
\end{equation}

\hfill 

Since $( J_d  Z_{k-1} - Z_{k-1}^\top  J_d ) \in \mathfrak{so}(3)$, equation~\eqref{eq: discrete EL 1} can be rewritten as $ S(a) = Z_k J_d  - J_d Z_k^\top$ with \begin{equation} a = h^2 \xi_k + h  \mathcal{f}_{d_k}^{R-} + h  \mathcal{f}_{d_{k-1}}^{R+}  + S^{-1}(J_d  Z_{k-1} - Z_{k-1}^\top  J_d).\end{equation} Given $(x_{k-1}, x_k, R_{k-1}, R_k, u_{k-1} , u_k)$, we first solve $ S(a) = Z J_d  - J_d Z^\top$ for $Z=Z_k$ as outlined in Remark~\ref{remark: Cayley}, and then get $R_{k+1} = R_k Z_k$. We then update $x_{k+1}$ using equation~\eqref{eq: discrete EL 2}. \\

\vspace{3mm}

 \subsection{Lie Group Forced Variational Integrator Networks (LieFVINs)}

 \hfill 
 \vspace{-3mm}

 We now describe the construction of Lie group Forced Variational Integrator Networks for the forced variational integrator on $\text{SE}(3)$ presented in Appendix~\ref{appendix: forced VI Lagrangian}, in the case where only position data is available. The LieFVIN is based on the discrete forced Euler--Lagrange equations~\eqref{eq: discrete EL 1}-\eqref{eq: discrete EL 3}. As before, the main idea is to parametrize the updates of the forced variational integrator and match them with the observed updates. \\

 We parametrize \textcolor{mygreen}{$m$}, $\mathcal{f}_d^{\pm}$ and $U$ as neural networks, and the matrix $J$ is a symmetric positive-definite matrix-valued function of $(x,R)$ constructed via a Cholesky decomposition $J = LL^\top $ for a lower-triangular matrix $L$ implemented as a neural network. We can also get $J_d = \frac{1}{2} \text{tr}(J) \mathbb{I}_3 - J$. To deal with the implicit nature of equation~\eqref{eq: discrete EL 1}, we propose two algorithms, based either on an explicit iterative solver or by penalizing deviations away from equation~\eqref{eq: discrete EL 1}:

\noindent \hrulefill
\vspace{1mm}

\noindent \textbf{Algorithm IIa.} Given $(\textcolor{mygreen}{x_0}, \textcolor{mygreen}{x_1}, R_0 , R_1, u_0, u_1) \mapsto (\textcolor{mygreen}{x_2},R_2) $ data, minimize discrepancies between the observed $ (\textcolor{mygreen}{x_2}, R_2)$ pairs and the predicted $(\textcolor{mygreen}{\tilde{x}_2}, \tilde{R}_2)$ pairs obtained as follows: \\

\noindent For each $(\textcolor{mygreen}{x_0}, \textcolor{mygreen}{x_1}, R_0,R_1, u_0, u_1)$ data tuple,
\begin{enumerate}
\item Get $\mathcal{f}^{R \pm}_{d_0}$, $\mathcal{f}^{R \pm}_{d_1}$, \textcolor{mygreen}{ $\mathcal{f}^{x \pm}_{d_0}$,$\mathcal{f}^{x \pm}_{d_1}$} from $(\textcolor{mygreen}{x_0}, \textcolor{mygreen}{x_1}, R_0,R_1, u_0, u_1)$, and  $ S(\xi_1) = \frac{\partial U_{1}}{\partial R_{1}}^\top R_{1} - R_{1}^\top \frac{\partial U_{1}}{\partial R_{1}}$
\item Get $\tilde{R}_2 = R_1 \text{Cay}(\mathcal{z})$ where $\mathcal{z}$ is obtained using a few steps of Newton's method to solve the vector equation~\eqref{eq: Vector Equation General Form} equivalent to 
\[ h^2 S(\xi_1) + h S( \mathcal{f}_{d_1}^{R-} + \mathcal{f}_{d_{0}}^{R+}  ) + ( J_d  Z_0 - Z_0^\top  J_d )   = Z J_d  - J_d Z \]   \vspace{-5.5mm}
\textcolor{mygreen}{ \item Compute $  \   \tilde{x}_2 = 2x_1 - x_{0} - \frac{h^2}{m}  \frac{\partial U_1}{\partial x_1} +  \frac{h}{m} R_{1} ( \mathcal{f}_{d_1}^{x-} +\mathcal{f}_{d_{0}}^{x} ) $}   \vspace{-0.5mm} \end{enumerate} 
\noindent \hrulefill

\newpage  

\noindent \hrulefill  \vspace{1mm}

\noindent \textbf{Algorithm IIb.}  Given $(\textcolor{mygreen}{x_0},\textcolor{mygreen}{x_1}, R_0 , R_1, u_0, u_1) \mapsto (\textcolor{mygreen}{x_2},R_2) $ data, minimize
\begin{itemize}
\item \textcolor{mygreen}{Discrepancies between observed $ x_2$ and predicted $\tilde{x}_2 =  2x_1 - x_{0} - \frac{h^2}{m}  \frac{\partial U_1}{\partial x_1} +  \frac{h}{m} R_{1} ( \mathcal{f}_{d_1}^{x-} +  \mathcal{f}_{d_{0}}^{x})  $}
    \item Deviations away from the equation 
    \begin{equation*}
    J_d (R_{0}^\top R_1 + R_{2}^\top R_1) - (R_1^\top R_{0}  + R_1^\top R_{2} )J_d   + h^2 \left( \frac{\partial U_{1}}{\partial R_{1}}^\top R_{1} - R_{1}^\top \frac{\partial U_{1}}{\partial R_{1}} \right) + h S( \mathcal{f}_{d_1}^{R-} + \mathcal{f}_{d_{0}}^{R+}  )  = 0
\end{equation*}  \vspace{-7mm} \end{itemize}
\noindent \hrulefill \vspace{3.5mm}

This general strategy extends to any other Lie group integrator. In particular, LieFVINs on $\text{SO}(3)$ can be obtained from the algorithms above as the special case where $x$ is constant, in which case we can disregard all the variables and operations in \textcolor{mygreen}{green}. Lie group variational integrator networks without forces (\textcolor{blue}{\textbf{LieVINs}}) can be obtained by setting $\mathcal{f}_{d_0}^{R\pm} = \mathcal{f}_{d_0}^{x\pm} = 0$. Note that the strategy behind Algorithm IIa enforces the structure of the system in a stronger way than in Algorithm IIb. However, for certain Lie groups and variational integrators, it might not be practical to use Newton's method to solve for the implicit updates, in which case Algorithm IIb is preferred. \\

When combined with MPC as described in Section~\ref{subsec:control_strategy}, the initial conditions $(q_{\ell -1}, q_\ell )$ for the optimal control problems can be obtained either from the position estimates $(\tilde{q}_{\ell -1} , \tilde{q}_\ell)$ or from (position,velocity) estimates $(\tilde{q}_\ell , \dot{\tilde{q}}_\ell)$ with finite difference approximations. As before, the Lie group constraints for the system do not need to be added as path constraints since they are automatically satisfied to (almost) machine precision, by the design of the LieFVINs. 

\end{document}